\documentclass{article}

\PassOptionsToPackage{numbers}{natbib}
\usepackage[preprint]{neurips_2024}

\usepackage[utf8]{inputenc} 
\usepackage[T1]{fontenc}    
\usepackage{hyperref}       
\usepackage{url}            
\usepackage{booktabs}       
\usepackage{amsfonts}       
\usepackage{nicefrac}       
\usepackage{microtype}      
\usepackage{xcolor}         
\usepackage{multirow}
\usepackage{caption}
\usepackage{makecell}
\usepackage{float}
\usepackage{graphicx}
\usepackage{subcaption}
\usepackage{changepage}
\usepackage{enumitem}

\title{LSM-MS2: A Foundation Model Bridging Spectral Identification and Biological Interpretation}

\author{
    \textbf{Gabriel Asher}\thanks{Equal contribution.} \quad
    \textbf{Devesh Shah}\footnotemark[1] \\
    \textbf{Amy A. Caudy} \quad
    \textbf{Luke Ferro} \quad
    \textbf{L\'{e}a Amar} \quad
    \textbf{Ana S. H. Costa} \quad
    \textbf{Thomas Patton} \\
    \textbf{Niall O'Connor} \quad
    \textbf{Jennifer M. Campbell} \quad
    \textbf{Jack Geremia}\\[0.5em]
    \textnormal{Matterworks, Inc.} \\
    \texttt{\{gabriel, devesh\}@matterworks.ai}
}

\begin{document}

\maketitle


\begin{abstract}
    A vast majority of mass spectrometry data remains uncharacterized, leaving much of its biological and chemical information untapped. Recent advances in machine learning have begun to address this gap, particularly for tasks such as spectral identification in tandem mass spectrometry data. Here, we present the latest generation of LSM-MS2, a large-scale deep learning foundation model trained on millions of spectra to learn a semantic chemical space. LSM-MS2 achieves state-of-the-art performance in spectral identification, improving on existing methods by 30\% in accuracy of identifying challenging isomeric compounds, yielding 42\% more correct identifications in complex biological samples, and maintaining robustness under low-concentration conditions. Furthermore, LSM-MS2 produces rich spectral embeddings that enable direct biological interpretation from minimal downstream data, successfully differentiating disease states and predicting clinical outcomes across diverse translational applications.


\end{abstract}

\section{Introduction}

Mass spectrometry coupled with liquid chromatography (LC–MS) offers a dense view of the molecular state of biological systems. This state captures metabolomic shifts invisible to other tools like sequencing or immunoassays \cite{ms_over_sequencing, ms_over_western_blots, ms_for_food_analysis}. Yet, the sparse, heterogeneous, and unstructured nature of LC–MS data limits its broader utility in scientific discovery \cite{ms_is_tough, dark_matter, dark_matter2}.

Machine learning (ML) on tandem mass spectrometry (MS/MS or MS2) provides a framework to analyze this high-dimensional and unstructured data\cite{ml_is_useful, ml_in_ms_review_paper}. Over the past decade, ML work has focused on improving chemical structure identification from MS/MS spectra (spectral identification) and demonstrating its effectiveness in annotating complex spectra \cite{MIST, MIST-CF, CFM-ID, ESP}. However, the broader potential of ML-driven MS2 models for biological applications---such as disease detection, metabolomic profiling, and mechanistic analysis---remains largely underexplored, especially when only limited task-specific data are available.

Foundation models offer a solution through learning rich, generalizable representations from large datasets, enabling multiple downstream tasks with minimal fine-tuning \cite{foundation_models_are_general, foundation_models_are_general2}. Analogous to natural language models that learn linguistic semantics \cite{llms_learn_semantics}, foundation models for MS2 data aim to learn the chemical semantics encoded in spectral patterns. Through training on large spectral datasets, these models can learn to generalize across instruments, molecules, and analytical tasks, often outperforming task-specific approaches. Previous foundation models for MS/MS \cite{casanova, dreams, iceberg} have laid important groundwork; here, we present the latest generation of our patented Large Spectral Model for MS2 (LSM-MS2)\cite{Matterworks2025PCT, gabriels_paper} and advance this line of work with two key contributions:

\begin{enumerate}[itemsep=0pt]
    \item Establish a new state of the art in spectral identification, emphasizing improved differentiation of key biologically relevant isomers from standard MS/MS spectra.
    \item Demonstrate that embeddings from the foundation model enable biologically meaningful interpretation, by-passing conventional MS workflows like feature detection and manual spectral annotation.
\end{enumerate}

\section{Background and Related Work}

Tandem mass spectrometry is a core analytical technique that provides molecular-level insights into a wide range of applications from drug discovery and clinical diagnostics to agriculture and environmental science \cite{noisy_data_and_ms_applications}. Each MS2 spectrum captures how a molecule fragments under controlled conditions, with precursor and fragment ions measured at single–part-per-million (ppm) mass accuracy \cite{mass_spec_book}.

Historically, MS2 data have been valuable in MS-based omics workflows which match experimental spectra to reference libraries of known compounds \cite{early_mass_spec}. This process underlies most conventional analysis pipelines such as those used in Compound Discover \cite{compound_discover}, Global Natural Products Social Molecular Networking (GNPS) \cite{GNPS}, and MZmine \cite{MZmine}. However, despite decades of progress, traditional library-based approaches have reached a practical ceiling. Recent estimates suggest that over 87\% of spectra in the GNPS repository remain unidentified despite rapid expansion of public spectral libraries \cite{unlabeledms287pct, dark_matter, vast_chemical_space}. This identification gap arises from many key factors including: library incompleteness and limited coverage of chemical space; experimental variability (e.g. collision energy and instrument type); fragmentation diversity from similar molecules; spectral leakage; and noisy peaks that obscure informative signals \cite{lack_of_chemical_diversity, experimental_variation, noisy_data_and_ms_applications}. 

To address these limitations, hybrid computational tools (e.g. SIRIUS \cite{SIRIUS}, MIST \cite{MIST}, MIST-CF \cite{MIST-CF}) combine rule-based fragmentation methods with machine learning to improve identification performance. These approaches often leverage probabilistic fragmentation trees alongside supervised graph neural networks or ensemble classifiers to predict molecular formulas or substructures. The rapid growth of these computational tools has also motivated the development of public benchmarks for systematic comparison, such as MassSpecGym \cite{MSG} and CASMI \cite{CASMI}.

For many scientists, analyte identification is only a single step toward answering broader biological questions. Researchers often acquire MS/MS data from small sample sets (fewer than a thousand) across diverse biological and experimental conditions (e.g., disease cohorts, time points, or clinical variables). This data is then used to uncover the molecular differences underlying these conditions. Running MS/MS on each sample can generate hundreds of thousands of spectra --- only a small fraction of which can be confidently annotated \cite{dark_matter}. Researchers then aim to draw biological conclusions from this sparse subset of labeled spectra to detect patterns or differences between experimental groups.

In recent years, several foundation models have been created to address the challenges of analyzing large-scale MS/MS datasets. Models such as Casanova \cite{casanova}, DreaMS \cite{dreams}, and ICEBERG \cite{iceberg} have set state-of-the-art performance on spectral identification and generative benchmarks, employing varied training approaches including self-supervised pretraining, autoregressive learning, and geometric deep learning. These methods combine large-scale learning techniques with domain knowledge to capture fragmentation patterns that generalize across chemical classes, instruments, and MS acquisition conditions. 

Among these, Casanova and DreaMS represent the closest precedents to our work. Casanova has been analyzed in the context of biological applications \cite{casanova_application}, however, its scope is limited to proteomics, which constrains its broader applicability to molecular omics. DreaMS, in contrast, focuses on small molecules and was trained on a chemical space similar to ours; yet it falls short of LSM-MS2 in spectral retrieval performance and does not demonstrate downstream biological interpretation. A follow-up study on ICEBERG \cite{iceberg2} explored the value of generative identification for biological problems, yet its explanatory power is restricted to a subset of predefined analytes of interest. While valuable for the final stages of biological discovery, these limitations motivate the development of models like LSM-MS2, which aim to provide versatile embeddings that can independently support both identification and direct biological interpretation without spectral annotation across diverse domains.
\section{Methods}
\label{sec:methods}
LSM-MS2 is a transformer-based foundation model trained on millions of MS/MS spectra. The training objective is designed to maximize separation in spectral space, producing a chemically meaningful embedding representation that generalizes across analytes, experimental conditions, and subproblems. In this paper we focus on evaluating only two primary domains of performance: spectral identification and biological interpretation. 

\subsection{Spectral Identification}
Spectral identification measures how well a model matches an experimental ("test") spectrum to those from known reference compounds. We compute a similarity score between each test spectrum and all reference embeddings in a curated library, rank the top matches, and quantify performance using standard retrieval metrics. To ensure a fair comparison and isolate LSM-MS2’s algorithm and training effects, all methods are evaluated using the same reference library and retrieval pipeline.

\subsubsection{Comparative Methods}
We benchmark LSM-MS2 against the following methods:

\begin{adjustwidth}{1em}{0pt}
\textit{Cosine Similarity:} The conventional approach in MS/MS identification, where the raw test spectrum is directly compared to each reference spectrum using cosine similarity. This approach forms the computational backbone of most non–machine-learning spectral matching tools. We use the MatchMS modified cosine implementation \cite{matchms}.

\vspace{0.5em}
\textit{DreaMS \cite{dreams}:} The current state-of-the-art deep learning model for spectral identification. We evaluate the DreaMS fine-tuned checkpoint \textit{embedding\_model.ckpt} \cite{Bushuiev2025DreaMS-ckpt} in an embedding-based retrieval setting, using cosine similarity between embedded spectra from the reference library.
\end{adjustwidth}

\subsubsection{Reference Library}
Our reference library comprises 1.8 million high-quality spectra corresponding to 99 thousand unique analytes. All entries were curated, quality-controlled, and merged across multiple public and internal sources. Full details of the library can be found in Appendix~\ref{appendix:dataset_details}.

\subsubsection{Benchmarking Datasets}
We evaluate spectral identification across three complementary datasets (full dataset and acquisition details are provided in Appendix~\ref{appendix:dataset_details}):

\begin{adjustwidth}{1em}{0pt}
\textit{MassSpecGym \cite{MSG}:} The most comprehensive public benchmark for MS/MS data, containing 231 thousand high-quality spectra spanning 29 thousand analytes. Minor data curation steps are detailed in Section \ref{sec:msg}.

\vspace{0.5em}
\textit{MWX-Isomers (Internal Benchmark):} A targeted dataset of 61 biologically relevant isomers across 22 isomer groups, collected to assess isomeric discrimination for analytes underrepresented in MassSpecGym. Only constitutional isomers were included, since stereoisomers cannot be reliably distinguished using MS data.

\vspace{0.5em}
\textit{NIST Dilution Series:} A NIST SRM 1950 human plasma dilution series used to evaluate performance in a biologically complex medium, encompassing a wide dynamic concentration range and realistic signal-to-noise conditions. We collected 84 samples with RP and HILIC methods in positive and negative modes at 7 dilutions (1:10, 1:20, 1:30, 1:40, 1:80, 1:120, and 1:160). Additionally, Thermo Acquire-X was collected for each method and mode at a 1:10 dilution.
\end{adjustwidth}

\subsubsection{Retrieval Metrics}
We quantify performance using two complementary retrieval metrics:

\begin{adjustwidth}{1em}{0pt}
\textit{Top-K Accuracy (Acc.):} Measures the fraction of test spectra correctly identified within the top K library matches. Consistent with the definition of \textit{Hit Rate @ K} in MassSpecGym \cite{MSG}. Scores range from 0 to 1, with 1 indicating perfect retrieval.

\vspace{0.5em}
\textit{Top-K Maximum Common Edge Subgraph (MCES) Distance:} Measures structural similarity between predicted and ground-truth molecular graphs as an edit distance—the minimum number of edges to remove from both graphs to achieve isomorphism. Scores of 0 indicate identical structures, with higher values reflecting greater dissimilarity. Computed using myopicMCES (threshold=15).
\end{adjustwidth}

\subsection{Biological Interpretation}
To evaluate the utility of LSM-MS2 embeddings for biological applications, we focused on publicly available studies that provide both MS/MS data and accompanying metadata. For modeling, each spectrum within a file is first encoded using LSM-MS2. We then apply an aggregation strategy to combine these individual spectrum embeddings into a single sample-level embedding that represents the corresponding file, and thus, biological sample. These sample-level embeddings serve as input features for conventional machine learning models to answer the specific biological question associated with each dataset.

The composition of the datasets, the nature of the downstream tasks, the evaluation metrics, and success criteria are defined by the independent studies themselves. We provide detailed descriptions of these characteristics for each biological study in the corresponding sections below.

\section{Evaluating Spectral Identification}

Tandem mass spectrometry datasets are most commonly used for spectral identification. In this section, we evaluate this task across multiple benchmarks, comparing LSM-MS2 to current state-of-the-art methods. To highlight the algorithmic and training advantages of LSM-MS2, all methods are evaluated using the same reference library and retrieval pipeline (see Section \ref{sec:methods}).

\subsection{MassSpecGym}
\label{sec:msg}
We first evaluated our model on MassSpecGym. During pre-processing, we removed 5,272 spectral duplicates, leaving 225,832 spectra for evaluation. We also identified 272 spectra where the provided InChIKeys did not match those generated from the associated isomeric SMILES; these were recalculated to ensure consistency. Following MassSpecGym’s precedent, evaluation was performed on the resulting 28,923 analytes defined by unique 2D InChIKeys.

\begin{table}[!b]
    \caption{Retrieval results on the MassSpecGym benchmark. “Per Spectrum” metrics are calculated across all spectra in the test set, whereas “Per Analyte” metrics are obtained by averaging results across spectra of each analyte prior to aggregation. The maximum achievable accuracies, constrained by reference library coverage, are 0.785 (per spectrum) and 0.823 (per analyte)}
    \centering
    \fontsize{7.6}{9.1}\selectfont
    \begin{tabular}{lcccccc}
        \toprule
        & \multicolumn{3}{c}{\makecell{Per Spectrum\\($N$ = 225{,}832 spectra)}} & \multicolumn{3}{c}{\makecell{Per Analyte\\($N$ = 28{,}923 analytes)}} \\
        \cmidrule(lr){2-4} \cmidrule(lr){5-7}
         & Top-1 Acc. $\uparrow$ & Top-5 Acc. $\uparrow$ & Top-1 MCES $\downarrow$ & Top-1 Acc. $\uparrow$ &  Top-5 Acc. $\uparrow$ & Top-1 MCES $\downarrow$ \\
        \midrule
        Cosine Similarity & 0.725 & 0.768 & 3.47 & 0.795 & 0.815 & 2.69 \\
        DreaMS & 0.726 & 0.770 & 3.52 & 0.794 & 0.817 & 2.67 \\
        LSM-MS2 (Ours) & \textbf{0.739} & \textbf{0.774} & \textbf{3.31} & \textbf{0.804} &  \textbf{0.820} & \textbf{2.47} \\
        \bottomrule
    \end{tabular}
    \label{tab:msg_results}
\end{table}

Performance is reported on both a per-spectrum and per-analyte basis (Table~\ref{tab:msg_results}). A key consideration is that retrieval-based spectral identification is constrained by the content of the reference library. This truism limits the maximum achievable accuracy to 0.785 per spectrum and 0.823 per analyte. We define this gap between previous best-performing method and the library-constrained maximum accuracy as the "remaining possible annotations". In this analysis, LSM-MS2 achieves a Top-1 Spectral Accuracy of 0.739—corresponding to 94\% of the maximum achievable accuracy and representing a 2\% improvement over prior methods—establishing it as the new state of the art. This performance gain accounts for 22\% of the remaining possible annotations. Furthermore, LSM-MS2 achieves a lower Top-1 MCES Distance than prior methods, indicating that when false positives occur, it retrieves analytes that are more chemically similar to the ground truth — reflecting a better-learned chemical space.

A second critical factor in assessing the performance of spectral identification is the proper handling of false positives. In scoring-based methods, the goal is to select a threshold that distinguishes true positives from false positives, but this threshold can vary significantly between the score distributions in different retrieval methods. We thus evaluate the separation between the distributions of true and false positives on MassSpecGym using ROC curves, which plot the true positive rate (TPR) against the false positive rate (FPR) across all scoring thresholds. The resulting area under the curve (AUC) is 0.950, 0.965, and 0.972 for Cosine Similarity, DreaMS, and LSM-MS2, respectively, indicating that LSM-MS2 achieves the best separation between TPR and FPR. The full analysis and ROC curves are provided in Appendix~\ref{appendix:msg_scores}.

\subsection{MWX-Isomers}
\label{sec:isomers}
One of the major challenges in spectral identification is the differentiation of isomeric compounds. Although these compounds are chemically similar and share identical formulae, they can have distinct functional roles and participate in different biological pathways. Isomers share an identical precursor mass and often produce highly similar fragmentation patterns. In conventional approaches, this large number of shared peaks among related compounds tends to saturate heuristic scoring methods, making it difficult to distinguish among isomers. A comparable limitation arises in machine learning models for MS2, where subtle intensity differences or fragment shifts between isomers often fall below the model’s discriminative threshold, leading to overlapping clusters in the learned embedding space.

Balanced performance across isomer groups is critical. It is a common misconception that achieving a top-1 accuracy of 1.0 on a single analyte is meaningful if its corresponding isomer has a much lower accuracy. Genuine isomeric discrimination requires that all unique members of an isomer group are correctly identified; otherwise, the model has not truly learned to separate the isomers.

\begin{figure}[!t]
  \centering
  \begin{subfigure}[t]{0.45\textwidth}
    \centering
    \includegraphics[width=\textwidth]{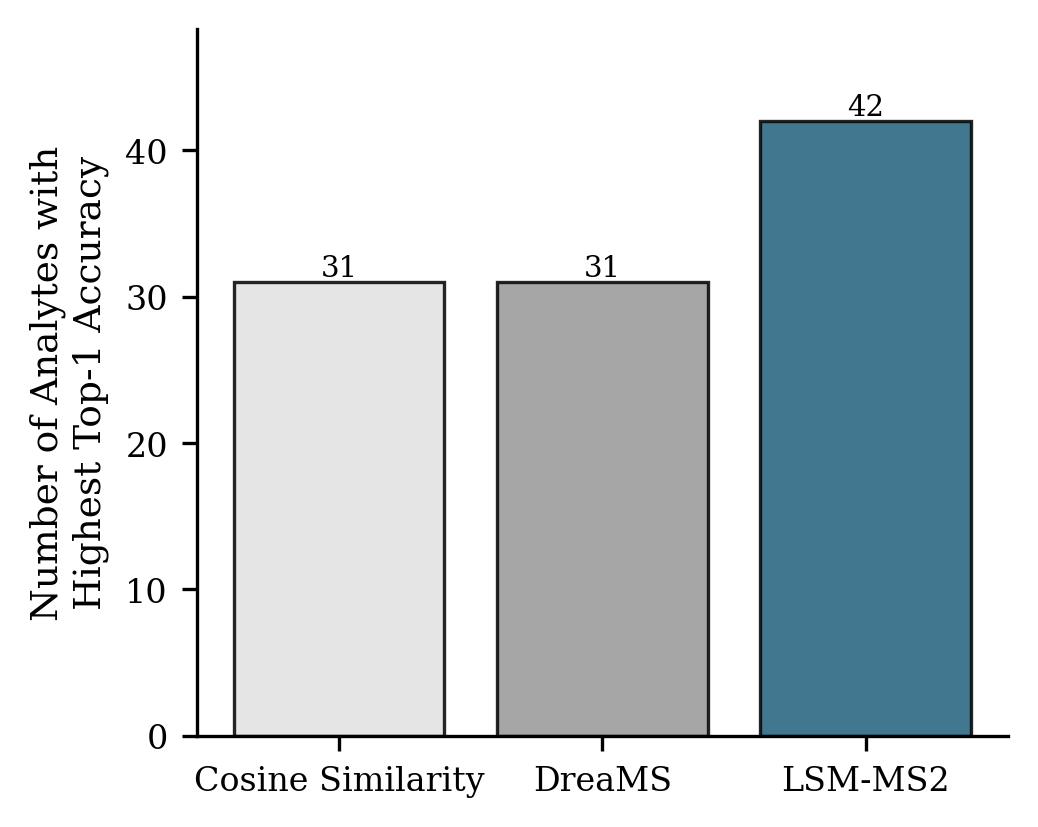}
    \caption{}
    \label{fig:all_isomers}
  \end{subfigure}
  \hfill
  \begin{subfigure}[t]{0.45\textwidth}
    \centering
    \includegraphics[width=\textwidth]{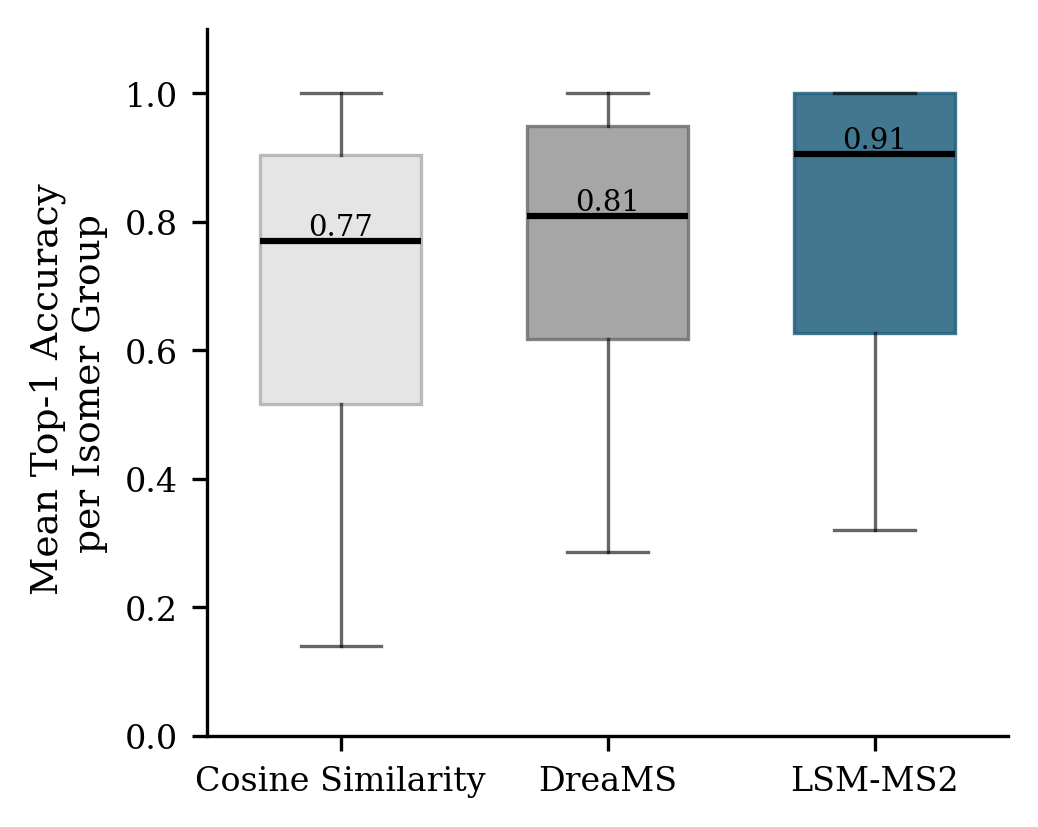}
    \caption{}
    \label{fig:isomers_boxplot}
  \end{subfigure}

  \caption{Comparison of model performance on the MWX-Isomers dataset. LSM-MS2 significantly outperforms previous methods on this dataset, in both a cumulative per-analyte stance, as well as per isomer groups. 
  (a) Overall per-analyte identification accuracy across all 61 biological isomers. In cases of tied top-1 accuracy for an analyte, each model achieving the tie receives one point, resulting in a cumulative total exceeding 61.
  (b) Per-group distribution of top-1 accuracies across all 22 isomeric groups in the dataset.}
  \label{fig:isomers_combined}
\end{figure}

We evaluate LSM-MS2 on a curated set of 61 biologically relevant isomers. Despite using no explicit isomer-focused contrastive supervision during training, LSM-MS2 outperforms previous state-of-the-art methods (Figure~\ref{fig:all_isomers}), correctly predicting nearly 30\% more analytes with higher top-1 accuracy than both Cosine Similarity and DreaMS. This analysis was conducted on the MWX-Isomers dataset, as unique exemplars of single isomers within the isomer groups of interest were either absent or represented by very few spectra in MassSpecGym. For completeness, detailed per-analyte results are provided in Appendix~\ref{appendix:isomers}, including performance on available MassSpecGym data.

In Figure~\ref{fig:isomers_boxplot}, we evaluate group-level isomeric performance by averaging top-1 accuracy across all analytes within each group. Among our 61 selected isomers, there are 22 isomer groups. LSM-MS2 outperforms other methods on these grouped metrics by 10\%, demonstrating consistent and balanced differentiation.

\begin{figure}[!t]
  \centering
  \includegraphics[width=1.0\textwidth]{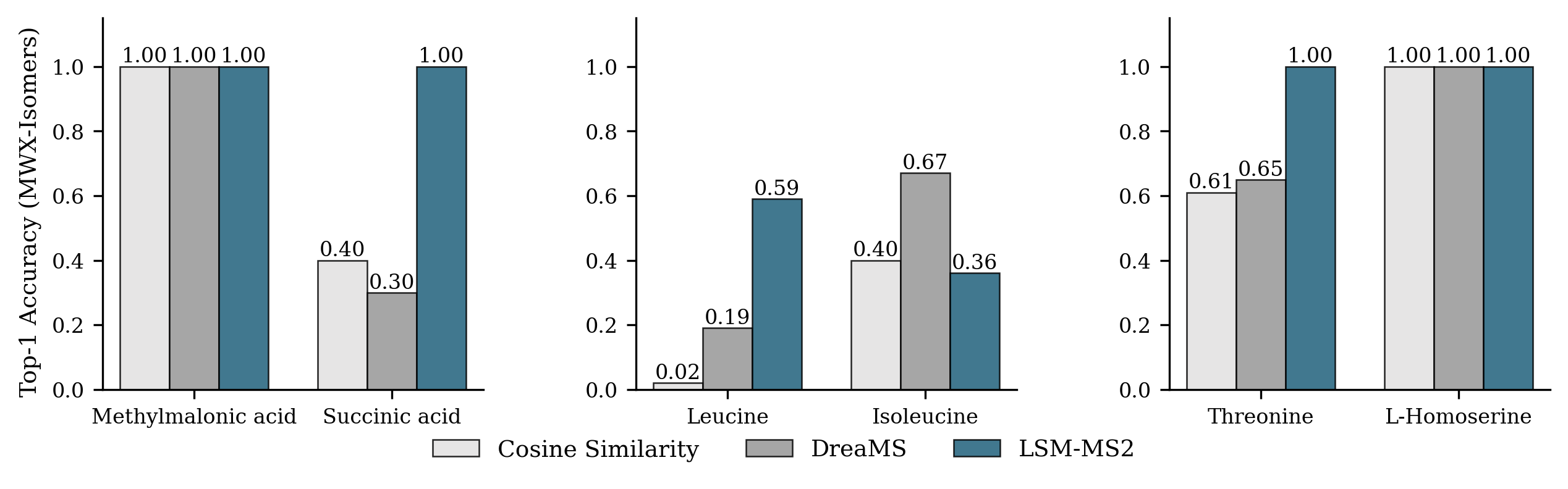}
  \caption{Top-1 identification accuracy across three biologically important isomer groups in the MWX-Isomers dataset. Balanced performance across all isomer pairs is critical for true isomeric discrimination, a task in which LSM-MS2 outperforms previous methods.}
  \label{fig:isomers_3_panel}
\end{figure}

Figure~\ref{fig:isomers_3_panel} highlights top-1 accuracy across three biologically important isomeric groups that are widely recognized as challenging. Across all three groups, LSM-MS2 exceeds the performance of previous state-of-the-art models. A particularly illustrative example is the isoleucine–leucine pair, a classic test of isomeric separation. While Cosine Similarity and DreaMS achieve higher accuracy on isoleucine, their performance on leucine is substantially lower, indicating asymmetric or biased classification. In contrast, LSM-MS2 achieves both higher and more balanced accuracy across the pair, yielding a mean top-1 accuracy of 0.48, reflecting improved discrimination.

\subsection{NIST Dilution Series}

We benchmarked identification performance on the NIST SRM 1950 human plasma reference material across a dilution series using LSM-MS2 and Cosine Similarity as implemented by \textit{MZmine} \cite{MZmine}, with configuration details provided in Appendix~\ref{appendix:nist}. MZmine hyperparameters were selected to maximize algorithmic comparability between methods rather than to individually optimize either approach.  An ablation study examining the impact of key MZmine hyperparameters is shown in Appendix~\ref{appendix:nist}. 

\begin{table}[!b]
    \caption{Global identification performance of Cosine Similarity vs LSM-MS2 at each method's optimal score threshold (Cosine = 0.90, LSM-MS2 = 0.89). "True Hit Rate" represents the fraction of correctly identified analytes relative to the maximum achievable hits based on the reference library. Precision and F1 Score are calculated using true positives (TP), spurious hits (FP), and false negatives (FN), where FN represents theoretically detectable analytes that were not identified.}
    \centering
    \resizebox{\linewidth}{!}{%
    \begin{tabular}{lccccc}
        \toprule
        & \multicolumn{2}{c}{\makecell{Counts}} & \multicolumn{3}{c}{\makecell{Performance Metrics (\%)}} \\
        \cmidrule(lr){2-3} \cmidrule(lr){4-6}
        & True Positives $\uparrow$ & Spurious Hits $\downarrow$ & Precision $\uparrow$ & True Hit Rate $\uparrow$ & F1 Score $\uparrow$ \\
        \midrule
        Cosine Similarity        & 125 & 390 & 24.3 & 28.2 & 26.1 \\
        LSM-MS2 (Ours) & \textbf{178} & \textbf{372} & \textbf{32.4} & \textbf{40.2} & \textbf{35.9} \\
        \midrule
        Relative $\Delta$ (\%) & \textbf{+42.4} & \textbf{--4.6} & \textbf{+33.3} & \textbf{+42.4} & \textbf{+37.5} \\
        \bottomrule
    \end{tabular}%
    }
    \label{tab:ms_id_performance}
\end{table}

In this dataset, ground-truth identification labels were defined as the subset of 443 analytes from \cite{Mandal2025SRM1950} present in our reference library. Any analyte identified by either method but absent from this subset was designated a \textit{spurious hit} and treated as a false positive for F1 score calculations. While treating spurious hits as false positives in F1 calculation is imperfect, since absence from the reference subset does not guarantee absence in the samples, it provides a consistent and comparable framework for evaluation. Retrieval performance was evaluated at three levels: globally across all samples, by dilution factor (aggregating data from all modes and LC methods within each dilution), and on a per-file basis.

\begin{figure}[!t]
  \centering
    \includegraphics[width=0.9\textwidth]{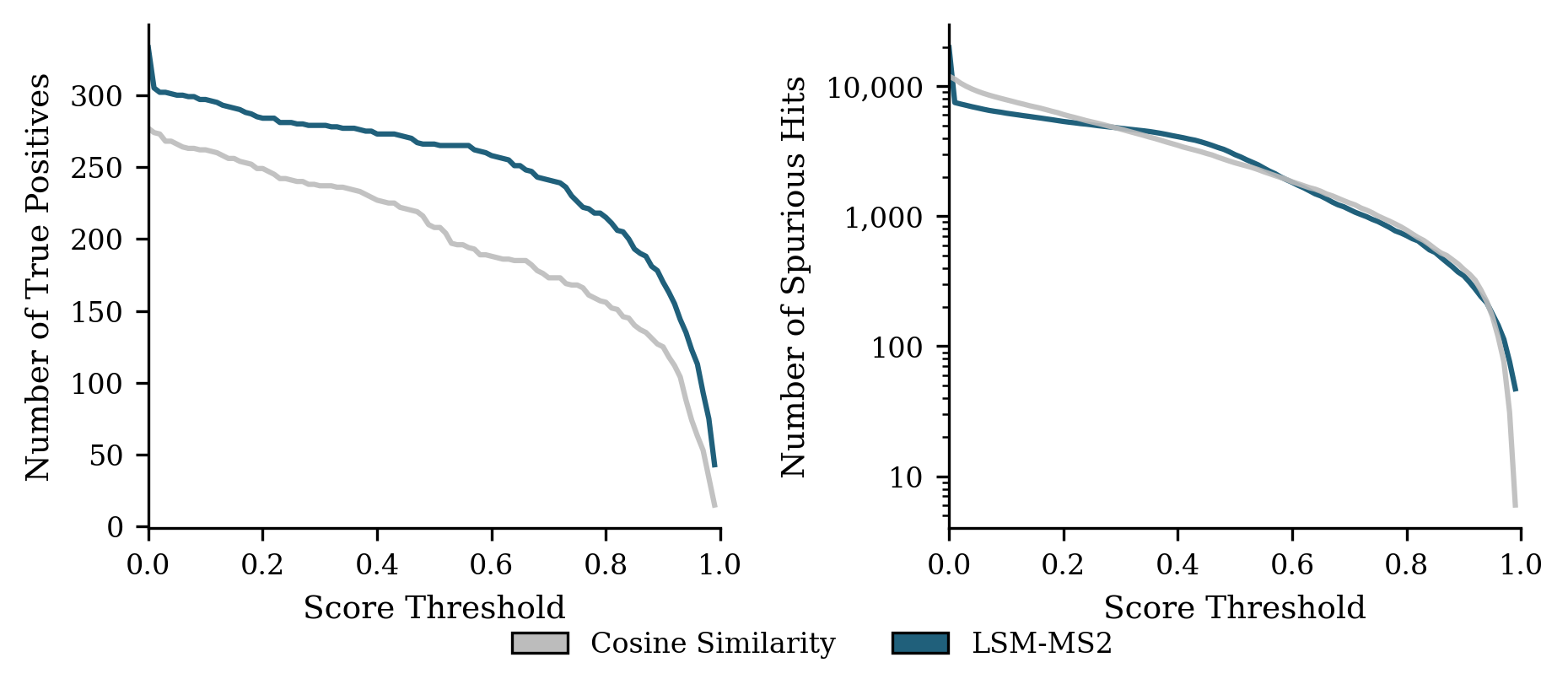}
  \centering
  \caption{Global identification performance of true positives (left) and spurious hits (right) comparing Cosine Similarity and LSM-MS2 across different score thresholds on the NIST Dilution Series.}
  \label{fig:2_panel_global}
\end{figure}

Figure~\ref{fig:2_panel_global} summarizes global performance across all scoring thresholds. LSM-MS2 consistently identifies more true positives and achieves higher precision than Cosine Similarity, a trend that persists even as Cosine Similarity is tuned to require more matched peaks (Appendix~\ref{appendix:nist}).To further quantify performance independently of score threshold, we evaluated each method at its optimal threshold---the score that maximizes F1. Table~\ref{tab:ms_id_performance} reports true positive counts, spurious hits, and derived performance metrics. At these thresholds, LSM-MS2 outperforms Cosine Similarity, retrieving 42.4\% more true positives and achieving 33.3\% higher precision, with no corresponding increase in spurious hits.

\begin{figure}[!t]
  \centering
    \includegraphics[width=0.9\textwidth]{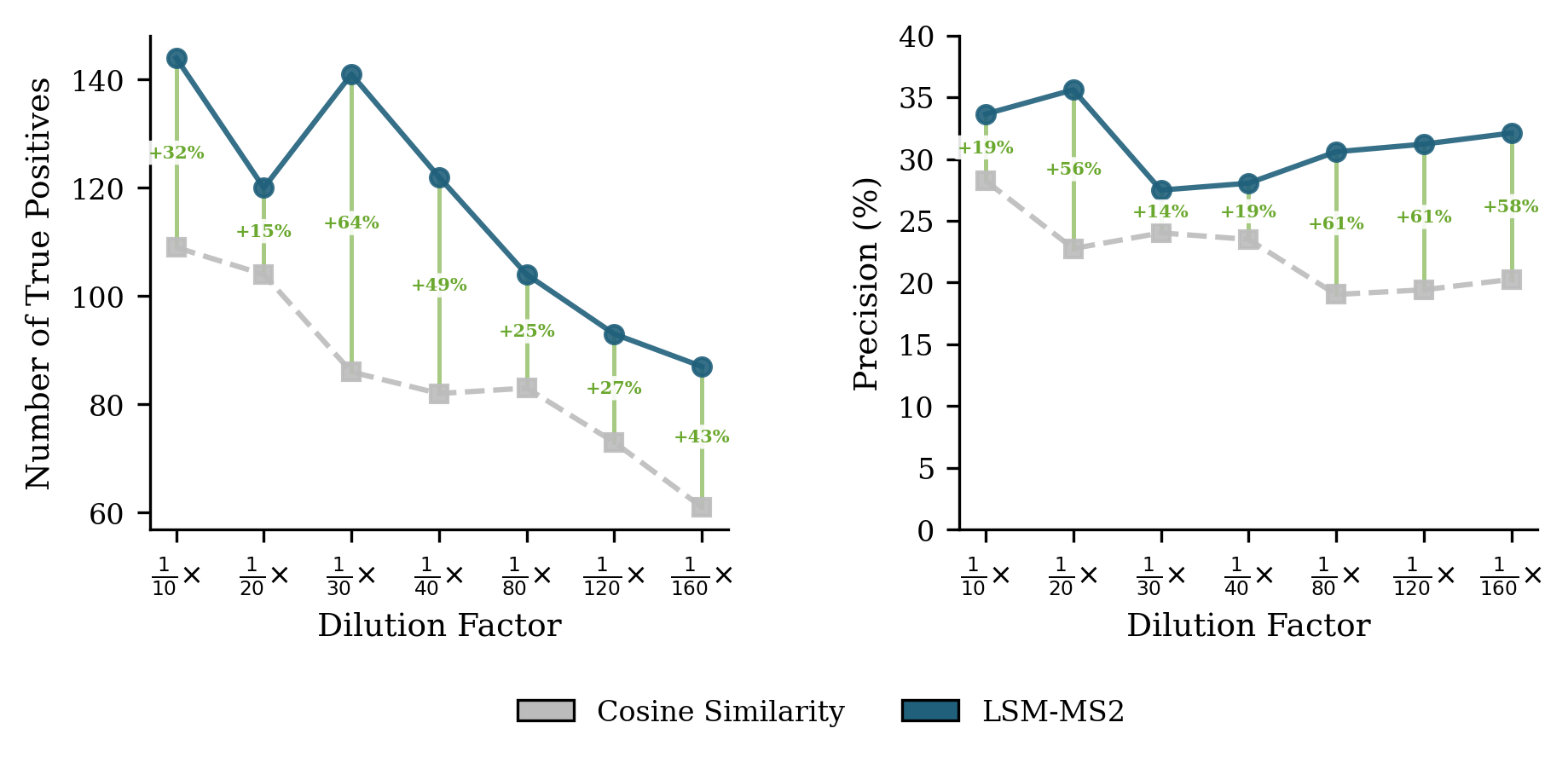}
  \centering
  \caption{True positive identifications (left) and precision (right) for both models at varying dilution factors. Metrics are evaluated at the score threshold that maximizes F1 for each scoring method-dilution pair.}
  \label{fig:dilution_optimal_f1}
\end{figure}

Performance in low-concentration regimes remains a persistent challenge for MS/MS-based identification - limiting both the number of analytes that can be identified from small sample amounts and the accuracy of detecting analytes present at endogenously low levels. To evaluate performance for analytes at low concentrations, we measured Cosine Similarity and LSM-MS2 performance at each dilution factor, using the score threshold that maximizes F1 for each scoring method and dilution pair. AcquireX data were excluded for this analysis. Figure \ref{fig:dilution_optimal_f1} compares the number of true positive identifications and precision across each dilution factor. While the number of true positives naturally decreases at higher dilutions, LSM-MS2 maintains consistent precision. Furthermore, the advantage of LSM-MS2 over Cosine Similarity holds as dilutions increase, indicating LSM-MS2’s robustness and effectiveness even under low-concentration conditions ($p < 0.001$, Appendix~\ref{appendix:nist}). Finally, on a per-sample basis, LSM-MS2 achieves higher F1 scores and identifies more true positives across all 84 samples, with higher precision in 90\% of cases. Complete results are provided in Appendix~\ref{appendix:nist}.

\section{Biological Interpretation}

Biological interpretation of mass spectrometry data is a complex and time-intensive process that often requires analytical method development to build reliable workflows that detect differential metabolic signals and interpret their underlying mechanisms. Accelerating this process remains a major challenge. Here, we demonstrate how LSM-MS2 enables rapid clustering and modeling of phenotypic endpoints, providing hypothesis testing in minutes rather than days or weeks.

\subsection{Antipsychotic Overdose Classification}
Fatal intoxication by antipsychotic agents remains a major challenge in forensic toxicology today. Using a dataset of 80 mouse plasma samples, \textit{Bai et al.}\cite{drug_fatality_study, mw_dataset_antipsychotics} performed LC–MS–based metabolomic profiling to investigate causes of death. The dataset comprised eight groups: four drug-induced fatalities (chlorpromazine (CPZ), perphenazine (PER), olanzapine (OLA), and clozapine (CLO)) and four non–drug-related controls (drowning, hemorrhagic shock, mechanical asphyxia, and cervical dislocation). The original study, which relied solely on identified analytes to separate cohorts, achieved clear separation between (1) overdose and control groups and (2) fatalities from CPZ versus OLA. However, it failed to differentiate CPZ from PER and OLA from CLO, attributed to their respectively shared pharmacodynamic receptor profiles \cite{mw_dataset_antipsychotics}.

\begin{figure}[!ht]
    \centering
    \includegraphics[width=1.0\linewidth]{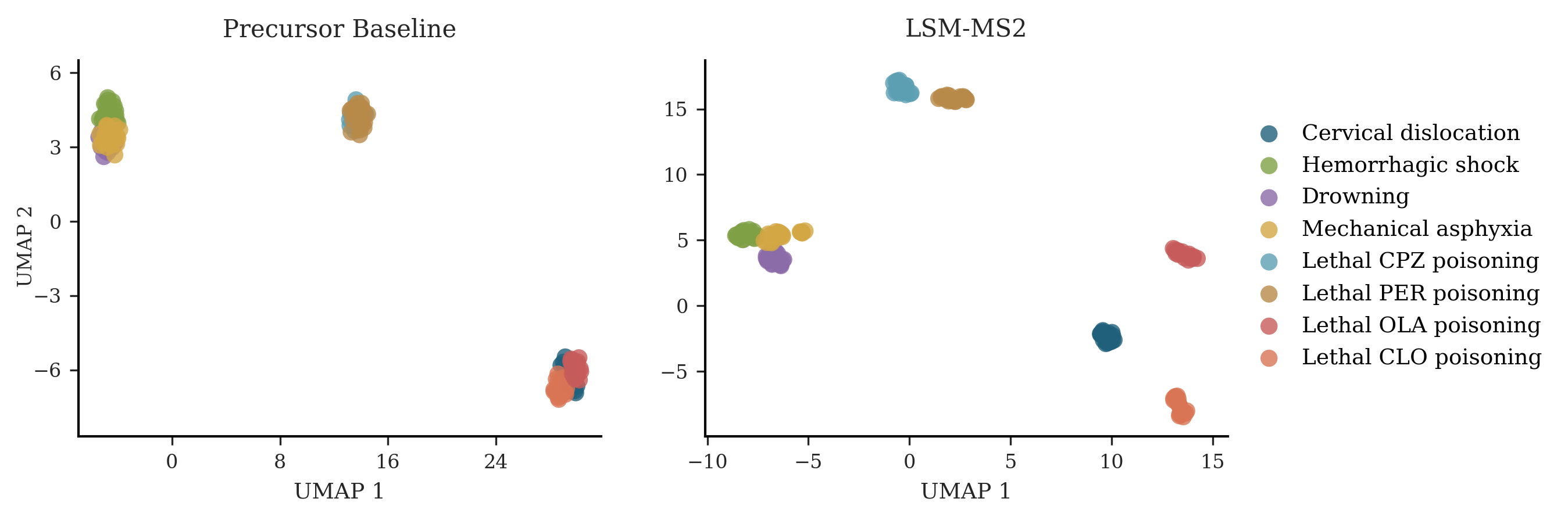}
    \caption{Unsupervised UMAP projections of study samples colored by mortality cause. Precursor baseline embeddings (left) show limited separation between CPZ/PER and OLA/CLO, whereas LSM-MS2 embeddings (right) reveal clear distinction across all fatality types.}
    \label{fig:drug_fatality_umap}
\end{figure}

To assess whether spectral representations can recover this missing structure, we construct a simple MS1 only (precursor) baseline embedding for comparison with LSM-MS2. Each sample is represented by a 0–1000~m/z binned vector, where the bin corresponding to each spectrum’s precursor m/z is incremented by 1, producing a coarse precursor–mass density profile matching LSM-MS2’s dynamic range. As shown in Figure~\ref{fig:drug_fatality_umap}, this baseline reproduces the same lack of separation reported by \textit{Bai et al.}

In contrast, clustering based on LSM-MS2 embeddings yields markedly improved resolution across all drug groups, successfully distinguishing CPZ/PER and OLA/CLO. Notably, samples corresponding to drowning, asphyxia, and hemorrhagic shock—conditions sharing hypoxic mechanisms—cluster closely together. These findings suggest that LSM-MS2 captures a more structured and biologically informative representation of metabolic variation than heuristic baselines.

\subsection{Septic Shock}
Sepsis is the life-threatening multiorgan dysfunction caused by a dysregulated response to infection, and can progress to septic shock, a state requiring intensive care. Septic shock has a mortality rate of 40 percent or higher as compared to a mortality rate of 10 percent for sepsis. Early recognition of septic shock is therefore critical for timely intervention and improved patient outcomes, particularly in the emergency department (ED). However, accurate diagnosis is challenging due to biological heterogeneity among pathogens, different infection sites, varied organ failure patterns, overlapping metabolic signatures from patient comorbidities, and most fundamentally the reality that septic shock is the extreme end of a continuum of septic states. 

\begin{figure}[!t]
    \centering
    \includegraphics[width=0.8\linewidth]{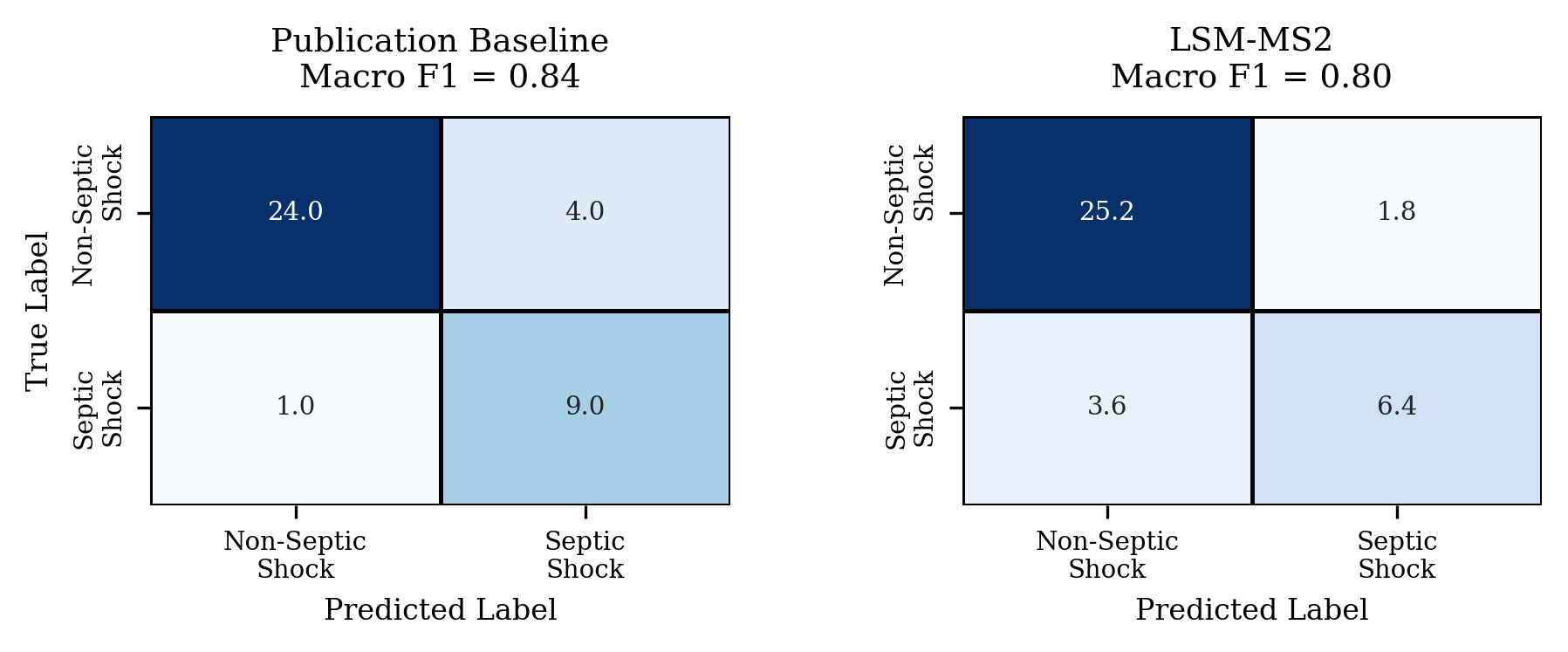}
    \caption{Confusion matrices for septic shock prediction. Results for LSM-MS2 are averaged over five random train/test splits.}
    \label{fig:septic_shock}
\end{figure}

A recent study by \textit{Hong et al.}\cite{septic_shock, mw_dataset_septic_shock} performed metabolomic profiling of serum collected at the time of admission from ED patients. Patients were selected for four cohorts to address the diagnostic challenges: uncomplicated sepsis without shock, patients who progressed to shock, patients with other types of shock, and patients admitted for other causes. The authors used the serum metabolomic profiling data to train a machine learning model on a curated panel of identified metabolites to predict early onset of septic shock with specificity and accuracy that exceeded existing clinical methods. Using the same dataset, we evaluate LSM-MS2 on this task to assess its translational relevance. We follow the original study’s 70/30 randomly stratified train–test split, average results over five seeds for reproducibility, and use the macro F1 score to account for class imbalance. Without completing any processes related to spectral identification, LSM-MS2 achieves a macro F1 of 0.80, closely matching the 0.84 reported in the original study seen in Figure ~\ref{fig:septic_shock} with total analysis time under one hour as compared to the time required for data processing and modelling. 

\subsection{Cystic Fibrosis}
\begin{figure}[!b]
    \centering
    \includegraphics[width=1.0\textwidth]{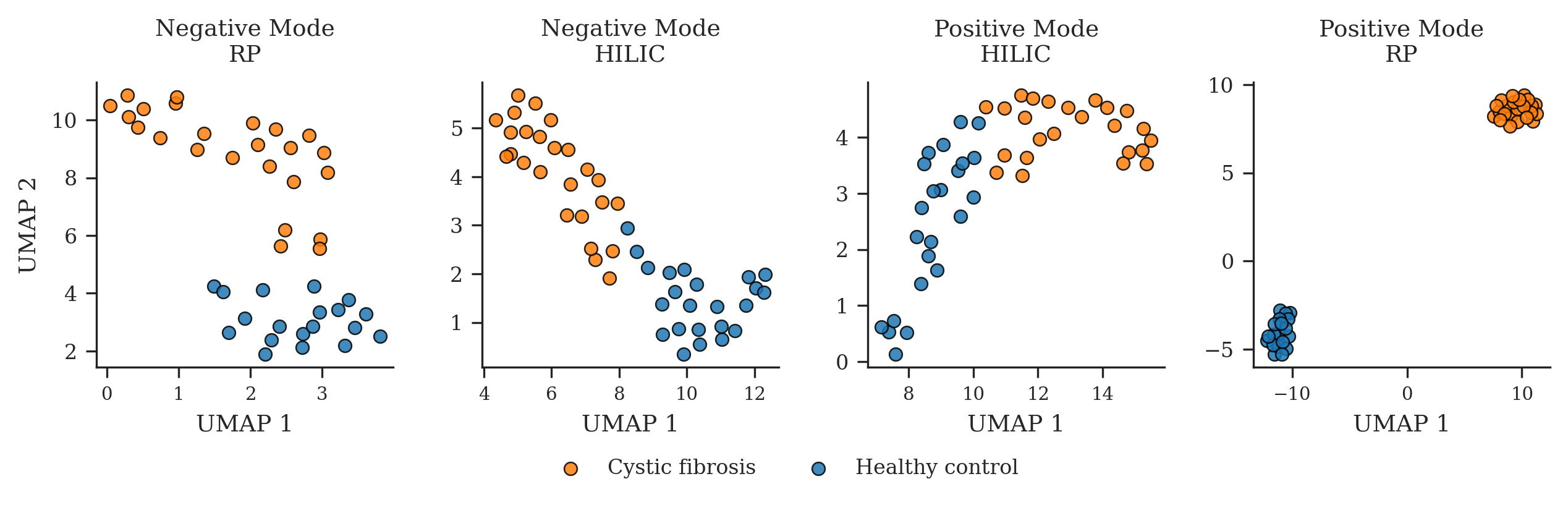}
    \caption{Unsupervised UMAP embeddings from LSM-MS2 of plasma samples across chromatography modes and ionization methods, colored by condition.}
    \label{fig:cystic_fibrosis}
\end{figure}
Cystic fibrosis (CF) is a multisystem genetic disorder characterized by chronic inflammation, oxidative stress, and metabolic dysregulation. In addition to the high risk of respiratory infections in CF patients, disruption in chloride ion channels impacts the function of other organ systems including pancreatic secretions that compromise nutrient uptake. A recent study \cite{mw_dataset_cystic_fibrosis} analyzed plasma from 24 CF patients and 26 age- and sex-matched healthy controls to investigate metabolomic alterations associated with CF. Samples were profiled using both reversed-phase (RP) and hydrophilic interaction (HILIC) liquid chromatography in positive and negative ionization modes. The combined metabolomic and lipidomic analysis revealed widespread metabolic disruptions underlying key aspects of CF pathophysiology.

We apply LSM-MS2 to assess which LC–MS configuration provides the strongest biological discrimination. As shown in Figure~\ref{fig:cystic_fibrosis}, unsupervised UMAP projections of LSM-MS2 sample embeddings reveal clear separation between CF and control cohorts across all chromatographic and ionization modes. Notably, the reversed-phase method in positive ionization mode yields the most pronounced distinction, indicating that it best captures disease-specific metabolic signatures or contains a key discriminating biomarker. This result illustrates how learned spectral representations can guide elements of experimental design and streamline method selection to enable the development of prognostic and diagnostic assays for CF.
\section{Conclusion \& Future Work}

In this work, we present the latest generation of LSM-MS2, a foundation model for MS/MS spectra that establishes a new standard for learned spectral embeddings. We outperform public benchmarks against previously published state-of-the-art models and demonstrate superior performance on additional metrics of key importance in the use of MS data in biological interpretation, including differentiation of isomers and identification of low-abundance compounds. Beyond spectral identification, we demonstrated that a well-learned embedding space can enable biologically meaningful insights across diverse studies and domains, even when only small datasets (or number of samples) are available.

Building on the promise demonstrated in this paper, two key directions remain to extend LSM-MS2 toward full molecular understanding from MS/MS data. First, to advance spectral identification, we aim to develop generative frameworks that infer molecular structures directly from spectra, reducing reliance on external libraries and enabling the discovery of truly novel compounds. Second, to enhance biological interpretation, we aim to improve the interpretability of LSM-MS2 embeddings to reveal how specific spectral features contribute to differences between sample groups, therein helping to translate patterns from small datasets into broader actionable biological insights. Together, these efforts will move LSM-MS2 beyond latent space separation to mechanistic insights linking spectral features to molecular and biological function.
\section{Acknowledgments and Disclosure of Funding}
We thank all members of Matterworks, Inc. for their advice and help on completing this work. All authors of this work are employees and shareholders of Matterworks, Inc.

\bibliographystyle{unsrtnat}


\newpage
\appendix
\section{Dataset Details}
\label{appendix:dataset_details}

In this section, we provide detailed descriptions of the datasets introduced in this work. To generate the internal datasets, we employed two primary liquid chromatography (LC) methods—Hydrophilic Interaction Liquid Chromatography (HILIC) and Reverse Phase (RP)—in both positive and negative ionization modes. Full details of the LC/MS methods are described here below.

\subsection{LC/MS Methods - HILIC Separation:}
A Transcend LX-2 multichannel system was used to inject 4 µL of sample into an Atlantis Premier BEH Z-HILIC VanGuard FIT Column (2.1 mm x 50 mm, 2.5 µm). Mobile phase A was composed of 20 mM ammonium carbonate in water with 0.25\% (v/v) ammonium hydroxide (pH 9.55). Mobile phase B was acetonitrile. All reagents were LC-MS grade. 
The elution gradient was as follows: 0-1 min, 95\% B; 1-8.5 min, ramp from 95\% to 20\% B; 8.5-9.5 min, 20\% B; 9.5-10 min, ramp from 20\% to 95\% B. The flow rate was 0.5 mL/min for the first 9.5 minutes and 0.8 mL/min thereafter. The autosampler was kept at 4 °C, and the analytical columns were held at 25°C. On the Orbitrap Exploris 120 and 240 mass spectrometers, MS1 data for each polarity was collected from 0 to 6.7 minutes at 60,000 resolution, and a scan range of 70 to 800 m/z. The RF lens was set to 55\%, the Automatic Gain Control (AGC) target was 1E6 ions with a maximum injection time of 60 ms.  The spray voltage was 3500 V and -2500 V in positive and negative ionization mode, respectively. The ionization source gas settings were the following: sheath gas 50, auxiliary gas 10, sweep gas 1 (arbitrary units). The ion transfer tube was kept at  315°C and the vaporizer temperature set to 350°C. Mild trapping and Run Start EASY-IC mode were enabled. Default charge state was 1 and the expected  peak width was 5 s. 

For Data Dependent Acquisition of MS/MS data an intensity threshold of 5E4 ions, an isolation window of 1.2 m/z, and apex detection of 30\% were selected. Four (Orbitrap Exploris 120) or twenty (Orbitrap Exploris 240) precursor per cycle were isolated within a 1.2 m/z window. Normalized stepped collision energies of 20, 50, and 100\% were applied. Fragment ions were scanned at 15,000 resolution.  

On the Orbitrap Astral mass spectrometer the RF lens was set to 50\%, the Automatic Gain Control (AGC) target was 1E6 ions with a maximum injection time of 50 ms. The spray voltage was 5500 V and -3500 V in positive and negative ionization mode, respectively. The ionization source gas settings were the following: sheath gas 40, auxiliary gas 8, sweep gas 1 (arbitrary units). Mild trapping was disabled. Advanced Peak Determination and Scan-to-Scan Start EASY-IC mode were enabled. Default charge state was 1 and the expected  peak width was 6 s. 

For Data Dependent Acquisition of MS/MS data on the Orbitrap Astral, an isolation window of 1.1 m/z was selected. The AGC target was 1000 ions and normalized stepped collision energies of 20\%, 50\%, and 100\% were applied. Fragment ions were scanned at 15,000 resolution.  

\subsection{LC/MS Methods - RP Separation:}
Each sample (4 µL) was injected into an ACQUITY UPLC HSS T3 column (2.1 mm x 50 mm, 1.8 µm) fitted with an ACQUITY UPLC HSS T3 VanGuard Pre-column (2.1 mm X 5 mm, 1.8 µm) using a Transcend LX-2 multichannel system. Mobile phase A was composed of 0.2\% formic acid in water (pH 2.5). Mobile phase B was 0.1\% formic acid in methanol. All reagents were LC-MS grade. 
Metabolites were eluted as follows: 0-0.8 min, 3\% B; 0.8-1 min, ramp to 40\% B; 1.8–2.3 min hold at 40\% B; 2.3–3.3 min, ramp to 70\% B; 3.30–4.30 min ramp from 70 to 90\% B; 4.3–4.8 min ramp from 90 to 95\% B; hold 95\% B for 2 minutes; return to 3\% B over the course of 0.6 minutes and hold from 7.6 to 10.8 min at 3\% B. The flow rate was 0.45 mL/min for the first 4.8 min, 0.5 mL/min from 4.8 to 6.8 minutes, 0.6 mL/min between 6.8 and 7.6 minutes, and finally 0.65 mL/min for the last 3.2 minutes. The autosampler was kept at 4 °C, and the column compartment was kept at 45°C. On the Orbitrap Exploris 120 and Exploris 240, the spray voltage was set to 3800 V and -2800 V in positive and negative ionization mode, respectively. The ionization source gases were set as follows for the first 4.8 minutes: sheath 50, auxiliary 10, sweep gas 1 (arbitrary units).  From 4.8 to 10.8 minutes the gas settings were: sheath 55, auxiliary 12, sweep gas 1 (arbitrary units). The ion transfer tube was kept at 335°C and vaporizer temperature 420°C. The MS1 data were collected from 0 to 7 minutes in each polarity, resolution 60,000, and a scan range of 75 to 950 m/z. The RF lens was set to 70\%, the maximum injection time to 60 ms and AGC target to 5E5 ions.  Mild trapping and Run Start EASY-IC mode were enabled. Default charge state was 1 and the expected  peak width was 4 s. 

For Data Dependent Acquisition of MS/MS data an intensity threshold of 5E3 ions, an isolation window of 1.2 m/z, and apex detection of 30\% were selected. Four (Orbitrap Exploris 120) or twenty (Orbitrap Exploris 240) precursor per cycle were isolated within a 1.2 m/z wimdow. Normalized stepped collision energies of 20\%, 50\%, and 100\% were applied. Fragment ions were scanned at 15,000 resolution.

\subsection{AcquireX Data Acquisition:}
For selected samples, AcquireX Deep Scan was used to create a background ion exclusion list from the extraction blank and ion inclusion list from the biological sample. The MS1 scans were acquired as described above. Seven injections of a representative sample were performed to generate MS2 spectra (ID files). For the ID files, monoisotopic precursor selection was enabled, the minimum intensity was 5000, charge states were filtered to 1, dynamic exclusion was set at auto, and target mass and targeted mass exclusions had a 5 ppm mass window. Twenty precursor ions per cycle were selected within a 1.0 Da isolation window and were fragmented by high energy collision-induced dissociation (30\%, 50\%, 150\% normalized stepped collision energy). MS2 fragment ions were scanned at 30,000 resolution with standard AGC target, maximum injection time of 54 ms, and 1 microscan.

\subsection{MWX-Isomers}
Spectra were acquired on ThermoFisher Scientific Orbitrap Exploris 240 and Astral instruments and for MS/MS spectra ramped collision energies [20, 50, 100] were used. Samples include both single-well injections and multi-analyte mixtures to capture varying spectral complexity.

\subsection{NIST SRM 1950 Human Plasma Dilution Series}
Metabolites in the NIST Standard Reference Material (SRM) 1950 (Metabolites in Frozen Human Plasma) were extracted fby precipitating proteins with an organic solution composed of 50\% methanol, 30\% acetonitrile, and 20\% water. For each sample type, multiple different sample to solvent ratios were used to create a dilution series.  

Data were acquired on a Thermo Fisher Orbitrap Exploris 240 using HILIC and RP chromatographic methods, in both positive and negative ionization modes, a described above.  The series comprised seven dilution levels (1:10, 1:20, 1:30, 1:40, 1:80, 1:120, and 1:160), with two technical replicates collected for each combination of dilution, liquid chromatography method, and mode. Additionally, we ran AcquireX Deep Scan on a single 1:10 dilution sample for each chromatography and mode.

\subsection{Reference Library}
Our reference library consists of approximately 1.8 million high-quality MS/MS spectra corresponding to over 99,000 unique analytes, aggregated from both public and internal sources. All spectra underwent rigorous quality control to ensure that only high-fidelity measurements were retained.

Public datasets incorporated into the reference library include: NIST23 \cite{NIST}, MS\textsuperscript{n}Lib \cite{msnlib}, GNPS \cite{GNPS-data}, MoNA \cite{MoNA2022}, and MassBank \cite{MassBank}.

Internal datasets include spectra from a curated collection of metabolomic analytes with biological and pharmacological relevance, acquired using the same LC/MS parameters described above.

Because MS/MS spectra alone cannot distinguish stereoisomers, all analytes in our reference library are defined by their 2D molecular structure (first 14 characters of the InChIKey). Stereoisomers sharing the same 2D structure were merged into a single entry, corresponding to the analyte with the lowest PubChem CID.

\section{Score Distribution in MassSpecGym}
\label{appendix:msg_scores}

In MS2 based retrieval tasks, it is critical not only to quantify the number of true positive hits but also to understand how frequently false positives occur and how reliably they can be distinguished from correct matches. Ideally, a scoring method should provide a confidence measure such that low-confidence predictions can be filtered, reducing the risk of reporting incorrect identifications.

\begin{figure}[!ht]
  \centering
  \begin{subfigure}[t]{0.48\textwidth}
    \centering
    \includegraphics[width=\textwidth]{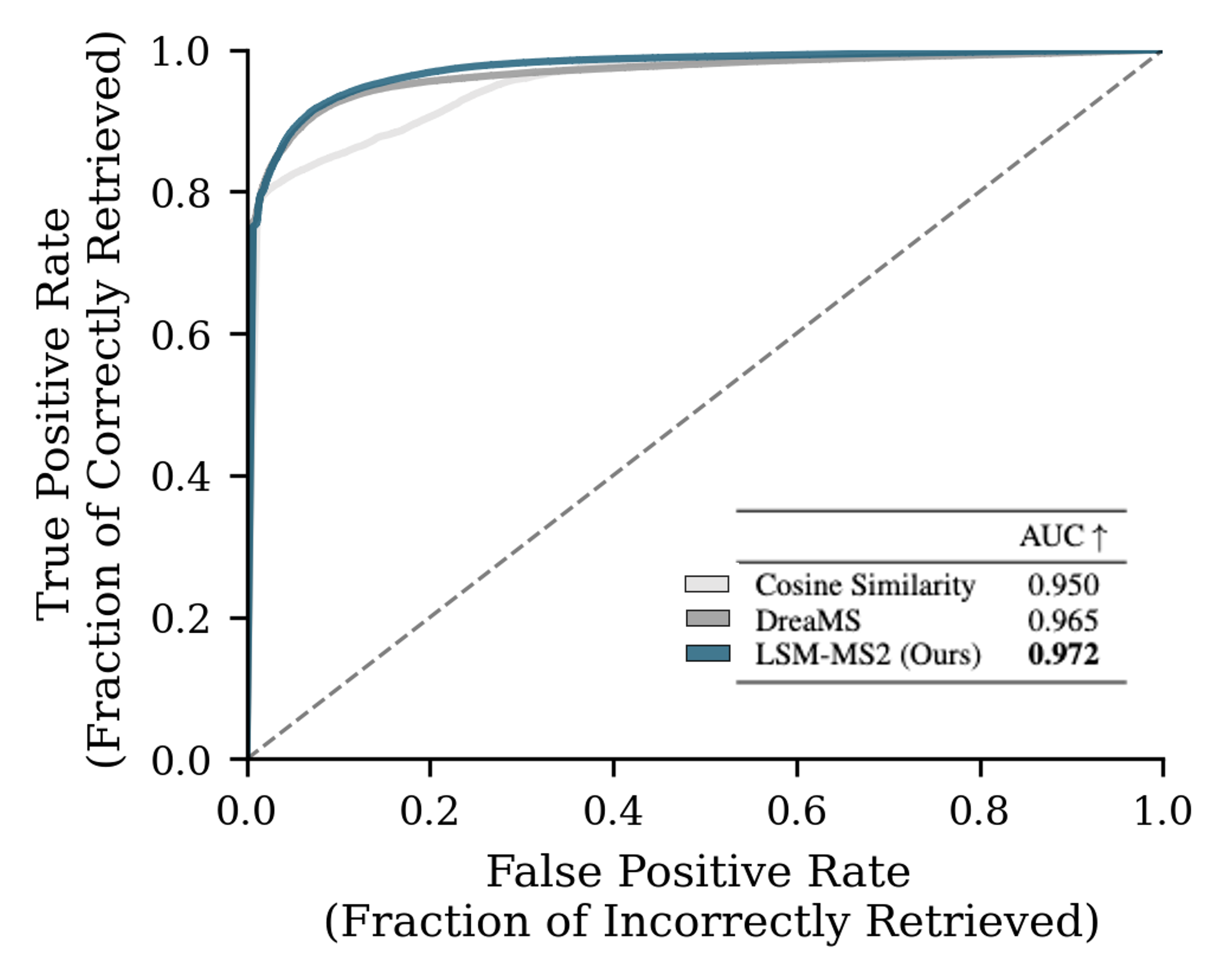}
    \caption{}
    \label{fig:msg_ROC}
  \end{subfigure}
  \hfill
  \begin{subfigure}[t]{0.48\textwidth}
    \centering
    \includegraphics[width=\textwidth]{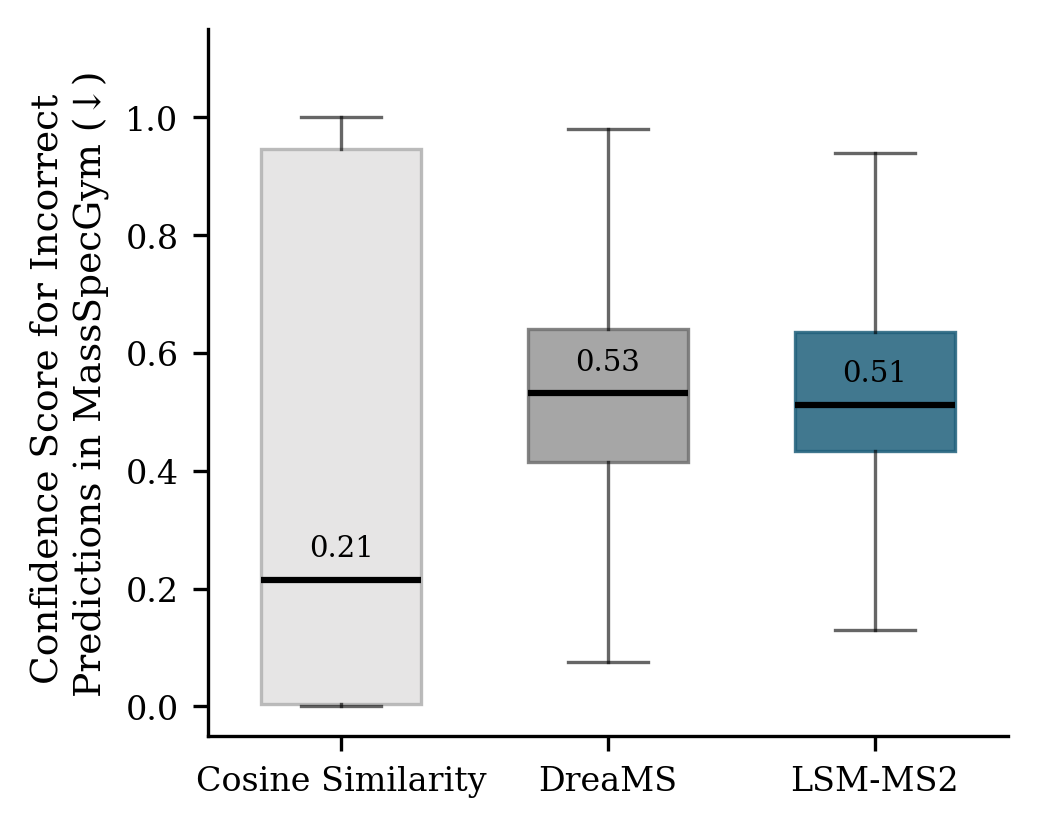}
    \caption{}
    \label{fig:msg_incorrect_score_boxplot}
  \end{subfigure}
  \caption{Score distributions on the MassSpecGym dataset. Only spectra with a potential precursor match in the reference library (not necessarily the correct analyte) are included, so all have an associated retrieval score. (a) ROC curves showing TPR vs FPR across all score thresholds. (b) Boxplots of false positive scores illustrate that LSM-MS2 better separates correct from incorrect matches compared to prior state-of-the-art methods.}
  \label{fig:msg_fpr}
\end{figure}

Here, we provide a detailed analysis of score distributions in MassSpecGym. For scoring-based methods, the key challenge is selecting a threshold that effectively separates true positives from false positives, which can vary across methods. To evaluate this separation, we employ ROC curves that plot the true positive rate (TPR) against the false positive rate (FPR) across all score thresholds. This approach enables a global comparison of methods independent of a single threshold choice.

Figure~\ref{fig:msg_ROC} shows that the area under the ROC curve (AUC) is 0.950 for Cosine Similarity, 0.965 for DreaMS, and 0.972 for LSM-MS2, indicating that LSM-MS2 achieves the best global separation between true and false positives.

\begin{figure}[!b]
  \centering
  \includegraphics[width=1.0\textwidth]{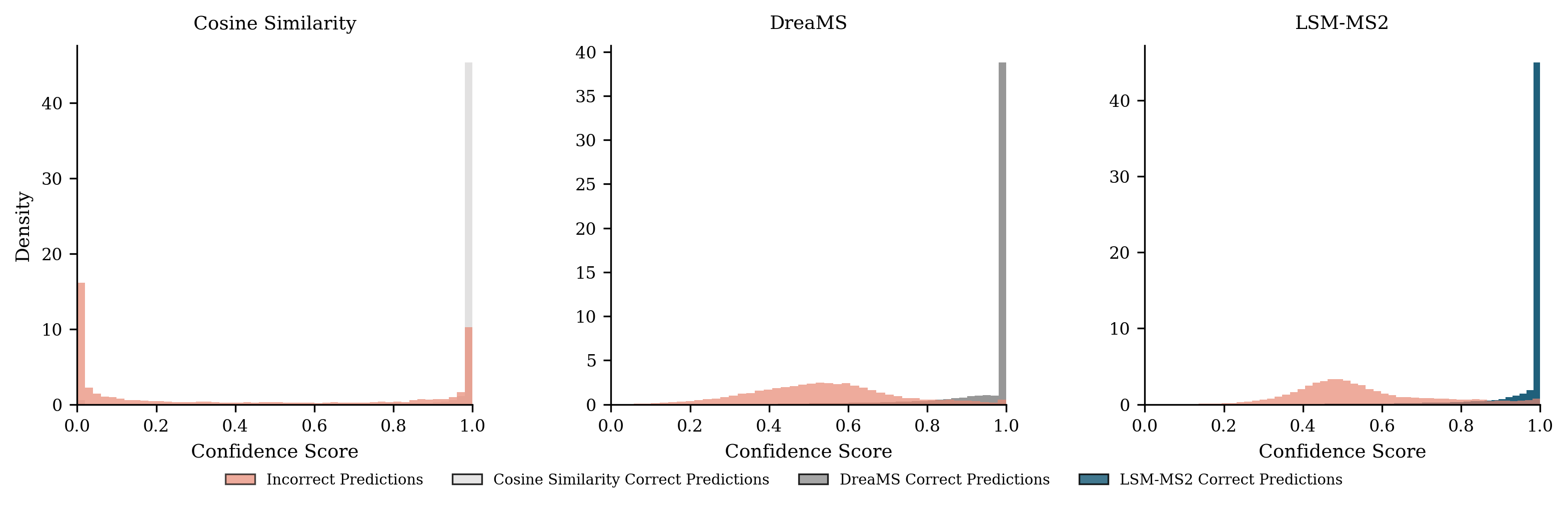}
  \caption{Histogram of true positive vs. false positive scores across all MassSpecGym samples, showing the differentiation between correct and incorrect matches for each method.}
  \label{fig:msg_score_histograms}
\end{figure}

We also examine the score distributions of false positives for each method (Figure~\ref{fig:msg_incorrect_score_boxplot}). While Cosine Similarity shows a lower median score for false positives, the distribution is broad, with over 25\% of false positives scoring above 0.9. Both DreaMS and LSM-MS2 have more concentrated distributions of false positive scores, allowing more reliable filtering. Note that showing only false positive scores provides limited insight; the critical measure is the separation between true positive and false positive score distributions. In this dataset, more than 95\% of true positives have scores above 0.95 for all three methods, making a direct overlay less informative. As a result, Figure~\ref{fig:msg_incorrect_score_boxplot} only highlights false positive distributions.

Figure~\ref{fig:msg_score_histograms} illustrates the differentiation between true and false positive score distributions. Cosine Similarity exhibits a bimodal false positive distribution, with scores clustering near 0 or 1, reflecting its tendency to either strongly reject or strongly accept a match regardless of correctness. In contrast, DreaMS and LSM-MS2 display smoother false positive distributions that are more concentrated at intermediate score values, providing a more graded measure of confidence and a clearer separation from true positive scores.

\begin{figure}[!ht]
  \centering
  \includegraphics[width=0.5\textwidth]{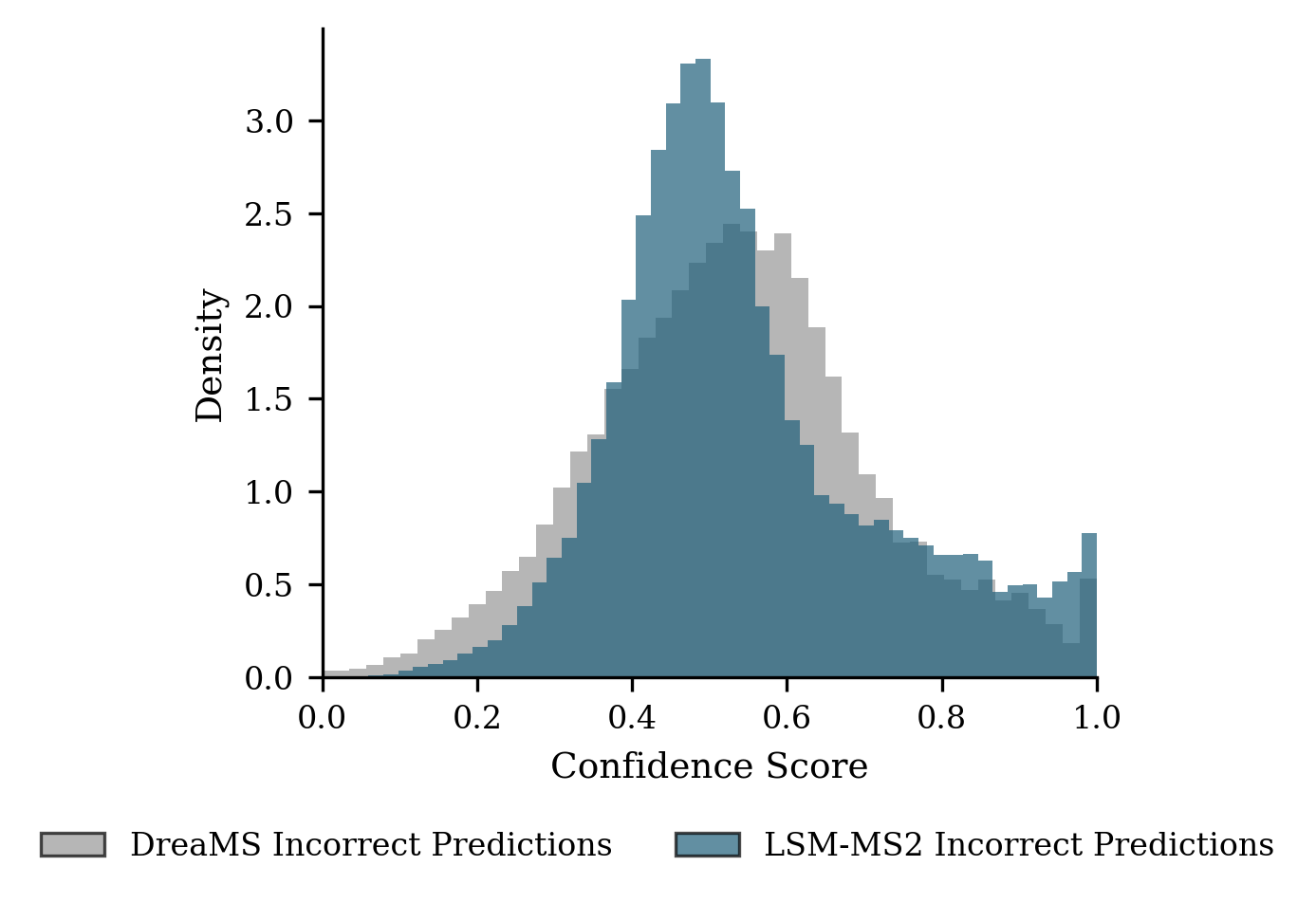}
  \caption{Comparison of false positive score distributions between DreaMS and LSM-MS2. True positive distributions are omitted as both methods show similar high-confidence scores ($>0.90$ for 95\% of true positives).}
  \label{fig:msg_incorrect_DreaMS_vs_LSM}
\end{figure}

To further compare DreaMS and LSM-MS2, Figure~\ref{fig:msg_incorrect_DreaMS_vs_LSM} focuses on false positive score distributions. Both methods produce false positive distributions that are relatively similar to each other, and in comparison to the true positive distributions, show a clear shift toward lower scores.
\section{Detailed Isomeric Spectral Identification Task}
\label{appendix:isomers}

Below, in Table~\ref{tab:isomers-full}, we present the complete results for all 61 biologically selected isomeric compounds examined in Section~\ref{sec:isomers}. Results are reported on two datasets: MWX-Isomers and MassSpecGym. The MWX-Isomers dataset was internally developed (and acquired) to enable systematic evaluation on these analytes, as MassSpecGym contains limited or no data for many of them, as shown below. Overall, LSM-MS2 consistently outperforms both Cosine Similarity and DreaMS in spectral identification across these challenging yet biologically meaningful cases. Note that all results use the same reference library for performance measurement, emphasizing that the improvements stem purely from algorithmic advances in LSM-MS2 that enable finer isomeric discrimination.

\begin{table}[!ht]
\caption{Full isomer discrimination performance across different methods}
\label{tab:isomers-full}
\centering
\fontsize{7.6}{9.1}\selectfont
\setlength{\tabcolsep}{3pt}
\begin{tabular}{@{}lrcccccccc@{}}
\toprule
&  & \multicolumn{4}{c}{MWX-Isomers} & \multicolumn{4}{c}{MassSpecGym} \\
\cmidrule(lr){3-6} \cmidrule(lr){7-10}
Analyte & \makecell{mono\\mass} & \makecell{$n$\\samples} & \makecell{Cosine\\Similarity} & DreaMS & LSM-MS2 & \makecell{$n$\\samples} & \makecell{Cosine\\Similarity} & DreaMS & LSM-MS2 \\
\midrule
beta-Alanine & 89.048 & 204 & 0.12 & 0.12 & \textbf{0.53} & 16 & \textbf{0.81} & \textbf{0.81} & \textbf{0.81} \\
Alanine & 89.048 & 204 & 0.33 & \textbf{0.48} & 0.25 & 25 & 0.72 & \textbf{0.92} & 0.76 \\
Sarcosine & 89.048 & 200 & \textbf{0.56} & 0.38 & 0.18 & 20 & \textbf{0.95} & 0.85 & 0.90 \\
\addlinespace[0.5em]
2-Aminoisobutyric acid & 103.063 & 16 & 0.12 & 0.50 & \textbf{0.62} & 40 & 0.88 & 0.85 & \textbf{0.90} \\
3-Aminoisobutyric acid & 103.063 & 20 & 0.40 & \textbf{0.70} & 0.30 & 11 & \textbf{1.00} & 0.82 & 0.82 \\
gamma-Aminobutyric acid & 103.063 & 214 & 0.47 & 0.50 & \textbf{0.51} & 46 & \textbf{0.91} & \textbf{0.91} & \textbf{0.91} \\
\addlinespace[0.5em]
2-Hydroxybutyric acid & 104.047 & 54 & \textbf{0.93} & 0.78 & \textbf{0.93} & 0 & -  & -  & -  \\
3-Hydroxybutyric acid & 104.047 & 22 & \textbf{1.00} & \textbf{1.00} & \textbf{1.00} & 6 & \textbf{1.00} & \textbf{1.00} & \textbf{1.00} \\
\addlinespace[0.5em]
Methylmalonic acid & 118.027 & 56 & \textbf{1.00} & \textbf{1.00} & \textbf{1.00} & 2 & \textbf{1.00} & \textbf{1.00} & \textbf{1.00} \\
Succinic acid & 118.027 & 208 & 0.40 & 0.30 & \textbf{1.00} & 0 & -  & -  & -  \\
\addlinespace[0.5em]
\bottomrule
\end{tabular}
\end{table}

\begin{table}[H]
\ContinuedFloat
\caption{Isomer discrimination performance across different methods (continued)}
\centering
\fontsize{7.6}{9.1}\selectfont
\setlength{\tabcolsep}{3pt}
\begin{tabular}{@{}lrcccccccc@{}}
\toprule
&  & \multicolumn{4}{c}{MWX-Isomers} & \multicolumn{4}{c}{MassSpecGym} \\
\cmidrule(lr){3-6} \cmidrule(lr){7-10}
Analyte & \makecell{mono\\mass} & \makecell{$n$\\samples} & \makecell{Cosine\\Similarity} & DreaMS & LSM-MS2 & \makecell{$n$\\samples} & \makecell{Cosine\\Similarity} & DreaMS & LSM-MS2 \\
\midrule
Threonine & 119.058 & 212 & 0.61 & 0.65 & \textbf{1.00} & 72 & \textbf{0.94} & \textbf{0.94} & \textbf{0.94} \\
L-homoserine & 119.058 & 8 & \textbf{1.00} & \textbf{1.00} & \textbf{1.00} & 36 & \textbf{0.94} & 0.86 & \textbf{0.94} \\
\addlinespace[0.5em]
Nicotinic acid & 123.032 & 38 & 0.47 & 0.79 & \textbf{1.00} & 74 & \textbf{0.99} & 0.97 & \textbf{0.99} \\
Picolinic acid & 123.032 & 18 & \textbf{1.00} & \textbf{1.00} & 0.89 & 31 & \textbf{1.00} & \textbf{1.00} & \textbf{1.00} \\
\addlinespace[0.5em]
Ketoisoleucine & 130.063 & 204 & 0.37 & 0.61 & \textbf{0.99} & 0 & -  & -  & -  \\
4-Methyl-2-oxopentanoic acid & 130.063 & 14 & \textbf{0.86} & 0.71 & 0.14 & 0 & -  & -  & -  \\
\addlinespace[0.5em]
Isoleucine & 131.095 & 212 & 0.40 & \textbf{0.67} & 0.36 & 118 & 0.78 & 0.81 & \textbf{0.84} \\
Leucine & 131.095 & 216 & 0.02 & 0.19 & \textbf{0.59} & 102 & \textbf{0.88} & 0.80 & 0.80 \\
L-Norleucine & 131.095 & 24 & 0.00 & 0.00 & \textbf{0.08} & 65 & \textbf{0.82} & \textbf{0.82} & 0.78 \\
\addlinespace[0.5em]
Dimethylmalonic acid & 132.042 & 2 & \textbf{1.00} & \textbf{1.00} & \textbf{1.00} & 0 & -  & -  & -  \\
Ethylmalonic acid & 132.042 & 42 & \textbf{1.00} & \textbf{1.00} & \textbf{1.00} & 0 & -  & -  & -  \\
Glutaric acid & 132.042 & 4 & 0.50 & \textbf{1.00} & \textbf{1.00} & 1 & \textbf{1.00} & \textbf{1.00} & \textbf{1.00} \\
Methylsuccinic acid & 132.042 & 36 & \textbf{1.00} & \textbf{1.00} & \textbf{1.00} & 3 & \textbf{1.00} & \textbf{1.00} & \textbf{1.00} \\
\addlinespace[0.5em]
Glycylglycine & 132.053 & 6 & \textbf{1.00} & \textbf{1.00} & \textbf{1.00} & 24 & \textbf{1.00} & \textbf{1.00} & \textbf{1.00} \\
Asparagine & 132.053 & 218 & 0.93 & 0.99 & \textbf{1.00} & 55 & \textbf{1.00} & \textbf{1.00} & \textbf{1.00} \\
\addlinespace[0.5em]
2,4-Dihydroxybenzaldehyde & 138.032 & 34 & 0.94 & 0.94 & \textbf{1.00} & 0 & -  & -  & -  \\
3-Hydroxybenzoic acid & 138.032 & 2 & \textbf{0.00} & \textbf{0.00} & \textbf{0.00} & 3 & \textbf{0.67} & \textbf{0.67} & \textbf{0.67} \\
4-Hydroxybenzoic acid & 138.032 & 34 & \textbf{0.94} & 0.88 & \textbf{0.94} & 22 & \textbf{1.00} & \textbf{1.00} & \textbf{1.00} \\
\addlinespace[0.5em]
2-Methylglutaric acid & 146.058 & 16 & 0.88 & \textbf{1.00} & 0.75 & 2 & \textbf{0.50} & \textbf{0.50} & \textbf{0.50} \\
3-Methylglutaric acid & 146.058 & 18 & \textbf{0.89} & 0.78 & 0.78 & 0 & -  & -  & -  \\
Adipic Acid & 146.058 & 12 & \textbf{0.83} & 0.50 & 0.33 & 5 & 0.80 & \textbf{1.00} & 0.80 \\
\addlinespace[0.5em]
Glutamic acid & 147.053 & 210 & 0.96 & 0.93 & \textbf{1.00} & 117 & \textbf{0.97} & \textbf{0.97} & \textbf{0.97} \\
N-Acetylserine & 147.053 & 18 & \textbf{0.78} & 0.33 & 0.44 & 17 & \textbf{0.94} & \textbf{0.94} & \textbf{0.94} \\
O-acetyl-L-serine & 147.053 & 18 & \textbf{1.00} & \textbf{1.00} & \textbf{1.00} & 34 & \textbf{0.97} & \textbf{0.97} & \textbf{0.97} \\
\addlinespace[0.5em]
2-Hydroxyglutaric acid (open) & 148.037 & 12 & \textbf{1.00} & \textbf{1.00} & \textbf{1.00} & 3 & \textbf{1.00} & \textbf{1.00} & \textbf{1.00} \\
3-Hydroxyglutaric acid & 148.037 & 10 & \textbf{1.00} & \textbf{1.00} & \textbf{1.00} & 1 & \textbf{1.00} & \textbf{1.00} & \textbf{1.00} \\
\addlinespace[0.5em]
2-Hydroxymethyl benzoic acid & 152.047 & 20 & 0.00 & \textbf{0.30} & 0.20 & 0 & -  & -  & -  \\
2-Hydroxyphenylacetic acid & 152.047 & 14 & \textbf{1.00} & 0.71 & 0.14 & 19 & \textbf{1.00} & 0.95 & 0.84 \\
2-Methoxybenzoic acid & 152.047 & 6 & \textbf{0.67} & 0.33 & 0.33 & 9 & 0.89 & \textbf{1.00} & 0.78 \\
3',5'-Dihydroxyacetophenone & 152.047 & 26 & \textbf{1.00} & 0.92 & 0.62 & 0 & -  & -  & -  \\
3-Methylsalicylic acid & 152.047 & 34 & \textbf{0.88} & 0.29 & 0.59 & 5 & \textbf{1.00} & \textbf{1.00} & \textbf{1.00} \\
4-Hydroxyphenylacetic acid & 152.047 & 6 & 0.67 & \textbf{1.00} & \textbf{1.00} & 11 & \textbf{1.00} & \textbf{1.00} & \textbf{1.00} \\
Phenoxyacetic acid & 152.047 & 20 & \textbf{0.00} & \textbf{0.00} & \textbf{0.00} & 0 & -  & -  & -  \\
Vanillin & 152.047 & 20 & 0.40 & \textbf{0.50} & 0.10 & 366 & 0.06 & 0.06 & \textbf{0.07} \\
p-Anisic acid & 152.047 & 18 & 0.56 & 0.56 & \textbf{0.67} & 3 & \textbf{1.00} & 0.67 & \textbf{1.00} \\
\addlinespace[0.5em]
Glycyl-L-proline & 172.085 & 28 & \textbf{1.00} & \textbf{1.00} & \textbf{1.00} & 12 & \textbf{1.00} & \textbf{1.00} & \textbf{1.00} \\
L-Prolylglycine & 172.085 & 20 & \textbf{1.00} & \textbf{1.00} & \textbf{1.00} & 3 & \textbf{1.00} & \textbf{1.00} & \textbf{1.00} \\
\addlinespace[0.5em]
Dehydroascorbic acid & 174.016 & 44 & \textbf{1.00} & \textbf{1.00} & 0.73 & 0 & -  & -  & -  \\
trans-Aconitic acid & 174.016 & 208 & 0.68 & 0.80 & \textbf{1.00} & 0 & -  & -  & -  \\
\addlinespace[0.5em]
Asymmetric dimethylarginine & 202.143 & 222 & 0.99 & 0.78 & \textbf{1.00} & 20 & \textbf{1.00} & \textbf{1.00} & \textbf{1.00} \\
Symmetric dimethylarginine & 202.143 & 26 & 0.00 & \textbf{0.92} & \textbf{0.92} & 9 & \textbf{1.00} & \textbf{1.00} & \textbf{1.00} \\
\addlinespace[0.5em]
Pseudouridine & 244.07 & 24 & \textbf{1.00} & \textbf{1.00} & \textbf{1.00} & 16 & \textbf{1.00} & \textbf{1.00} & \textbf{1.00} \\
Uridine & 244.07 & 218 & 0.99 & 0.98 & \textbf{1.00} & 78 & 0.55 & 0.55 & \textbf{0.56} \\
\addlinespace[0.5em]
Glucose 6-phosphate & 260.03 & 10 & \textbf{0.00} & \textbf{0.00} & \textbf{0.00} & 0 & -  & -  & -  \\
Fructose 6-phosphate & 260.03 & 212 & 0.19 & 0.52 & \textbf{0.94} & 8 & \textbf{0.62} & \textbf{0.62} & \textbf{0.62} \\
Galactose 1-phosphate  & 260.03 & 18 & \textbf{1.00} & \textbf{1.00} & \textbf{1.00} & 32 & \textbf{1.00} & 0.97 & \textbf{1.00} \\
\addlinespace[0.5em]
1-Methylguanosine & 297.107 & 28 & \textbf{1.00} & 0.79 & 0.79 & 11 & 0.91 & \textbf{1.00} & 0.91 \\
2-Methylguanosine & 297.107 & 34 & 0.35 & \textbf{0.59} & 0.35 & 17 & \textbf{0.94} & 0.82 & 0.76 \\
3'-O-Methylguanosine & 297.107 & 14 & 0.14 & \textbf{1.00} & \textbf{1.00} & 1 & 0.00 & 0.00 & \textbf{1.00} \\
\addlinespace[0.5em]
Adenosine monophosphate & 347.063 & 228 & 0.96 & 0.93 & \textbf{0.99} & 62 & \textbf{0.92} & 0.89 & 0.90 \\
Adenosine-2'-monophosphate & 347.063 & 18 & 0.67 & \textbf{1.00} & \textbf{1.00} & 0 & -  & -  & -  \\
\bottomrule
\end{tabular}
\end{table}

\section{NIST: Supplemental Analysis}
\label{appendix:nist}

\subsection{Statistical Testing of Precision at Low Concentration}

\begin{figure}[!ht]
    \centering
    \includegraphics[width=0.65\textwidth]{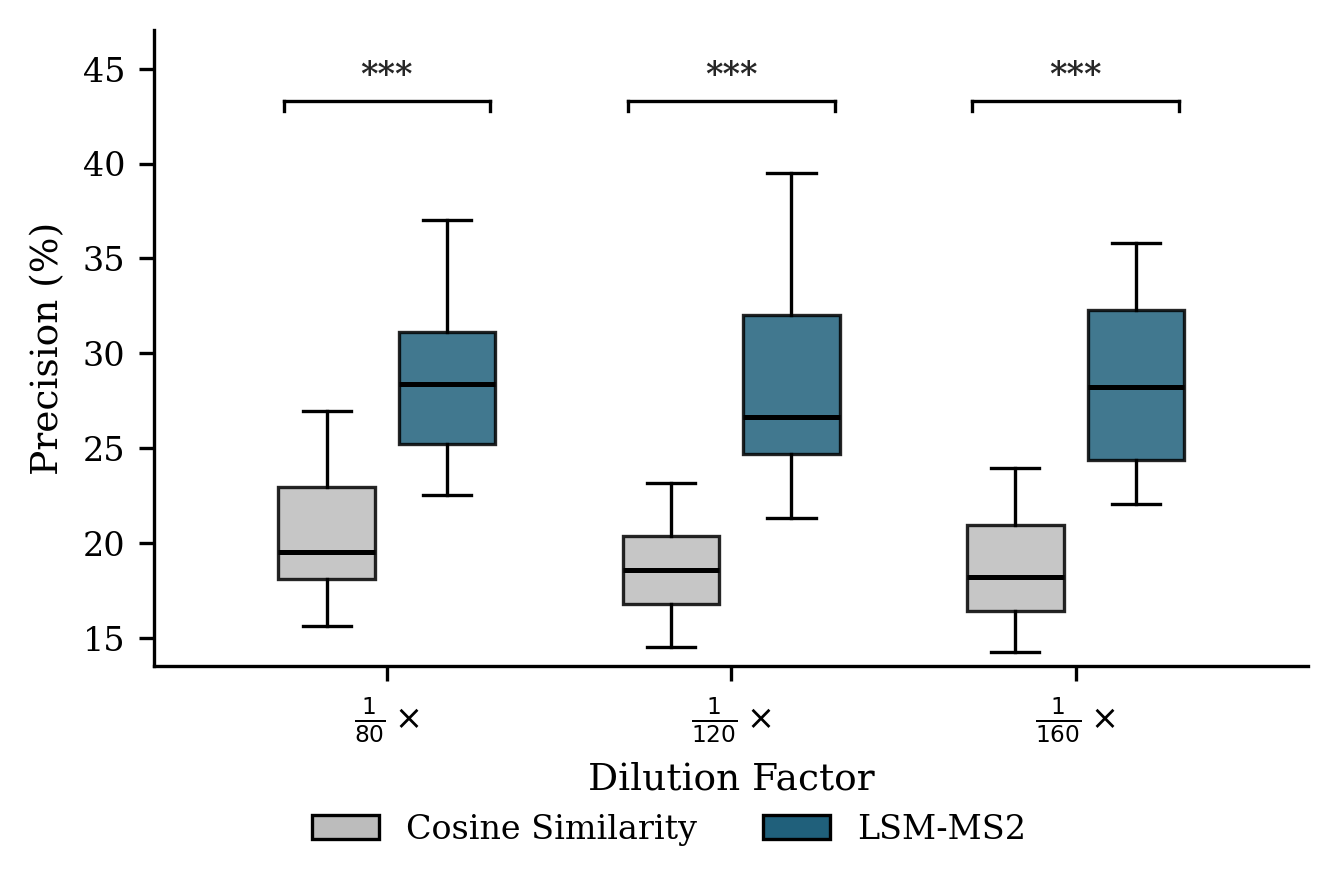}
    \caption{Statistical testing on low-concentration samples. A one-sided Welch’s t-test was applied to the top 20 F1 thresholds for each method and dilution to evaluate the claim that LSM-MS2 exhibits higher precision than cosine similarity at low-concentration. $^{***}$ denotes $p < 0.001$.}
    \label{appendix:statistical-testing}
\end{figure}

We report statistical testing on our claim that LSM-MS2 achieves higher precision than cosine similarity in low-concentration samples. To this end, we performed a one-tailed Welch’s t-test comparing cosine similarity and LSM-MS2 results using the top 20 F1 score thresholds between 0–1 for each dilution and model~Figure~\ref{appendix:statistical-testing} presents \. Calculated \textit{p}-values are 2.1$\times$10\textsuperscript{--8}, 2.0$\times$10\textsuperscript{--8}, and 1.1$\times$10\textsuperscript{--9} for the 1:80, 1:120, and 1:160 dilutions, respectively. These results demonstrate that LSM-MS2 achieves statistically significantly higher precision than cosine similarity at low concentrations.

\subsection{Per-File Comparison of Identification Performance}

We next evaluated LSM-MS2 and cosine similarity on a per-file basis to assess performance consistency across samples. Each value in Table~\ref{tab:head_to_head_filewise} represents the number of files where a given method achieved superior performance for the specified metric. For each sample and method, we determine the score threshold that maximizes the F1 score. Across all n=84 samples, LSM-MS2 achieves higher F1 scores and a greater number of true positives than Cosine Similarity. This improvement comes with minimal cost to precision: LSM-MS2 exhibits higher precision in 90\% of samples.
\begin{table}[!ht]
    \caption{Head-to-head comparison of per-file performance between cosine similarity and LSM-MS2. Each value represents the number of files in which the given method achieved superior performance for the specified metric.}
    \centering
    \resizebox{\linewidth}{!}{%
    \begin{tabular}{lccccc}
        \toprule
        & True Positives & Spurious Hits & Precision & True Hit Rate & F1 Score \\
        \midrule
        Cosine Similarity & 0 & 30 & 8 & 0 & 0 \\
        LSM-MS2 (Ours) & 84 & 54 & 76 & 84 & 84 \\
        \bottomrule
    \end{tabular}%
    }
    \label{tab:head_to_head_filewise}
\end{table}

\begin{figure}[!ht]
  \centering
  \includegraphics[width=1.0\textwidth]{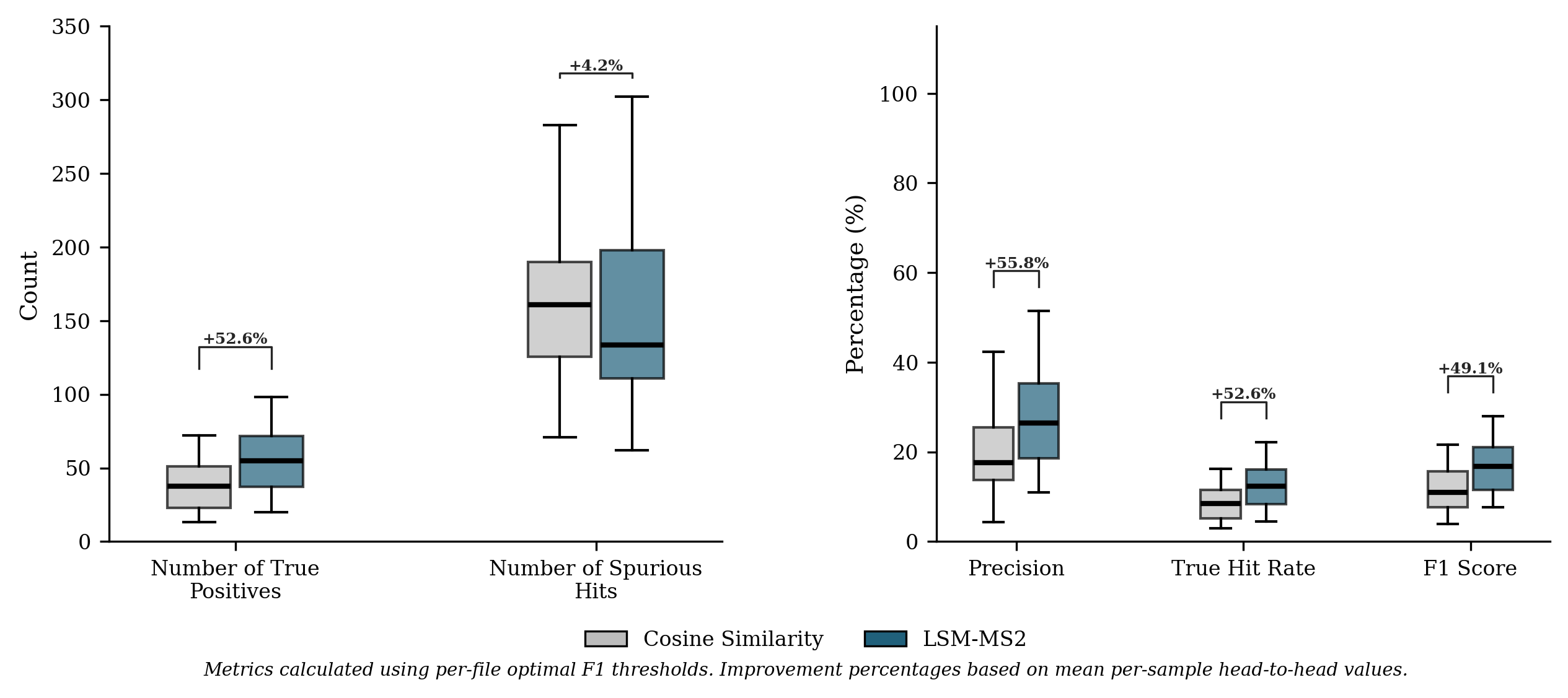}
  \caption{Boxplots of per-file performance comparing LSM-MS2 and cosine similarity. Values represent mean differences per file across metrics.}
  \label{appendix:per-sample}
\end{figure}

We compare Cosine Similarity and LSM-MS2 on a per-sample basis using quantitative performance metrics in Figure~\ref{appendix:per-sample}.LSM-MS2 exhibits 52.6\% and 55.8\% average improvements in true positives and precision relative to Cosine Similarity. However, these gains are accompanied by a slight increase in the mean number of spurious hits. While LSM-MS2 produces fewer spurious hits in 64\% of samples and a lower median count per sample, its mean spurious hit rate is 4.2\% higher than that of Cosine Similarity. We attribute this effect to a small subset of samples where the optimal LSM-MS2 score threshold occurs near zero, resulting in a few cases with substantially more spurious hits than Cosine Similarity.

\subsection{Robustness Analysis}

\begin{figure}[!ht]
  \centering
  \includegraphics[width=0.4\textwidth]{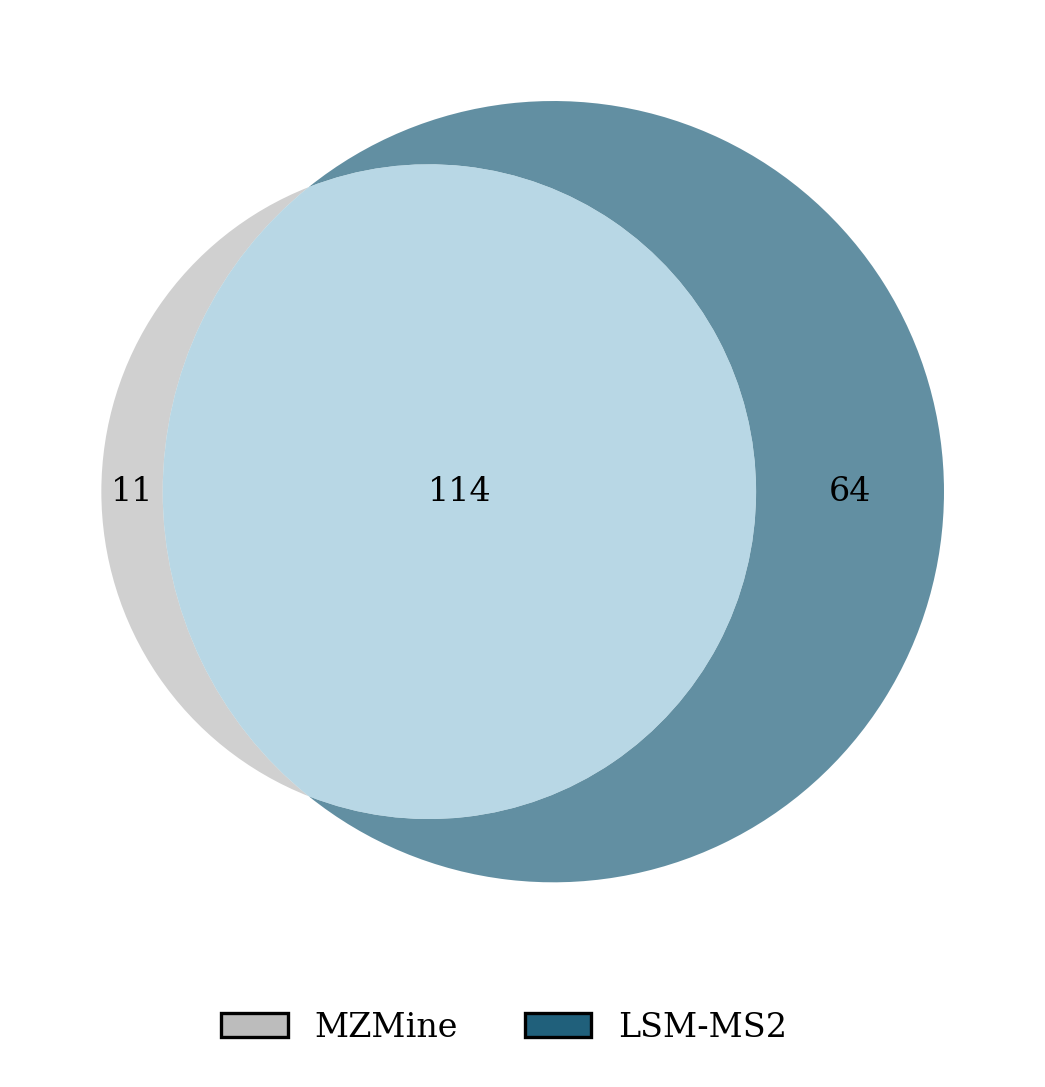}
  \caption{Venn diagram of true positive hits at the optimal F1 threshold between LSM-MS2 and MZmine.}
  \label{appendix:global-venn-diagram}
\end{figure}

To confirm that LSM-MS2 is building on the identifications of Cosine Similarity instead of identifying an orthogonal set of true positives, we plot a Venn diagram of global true positive hits between Cosine Similarity and LSM-MS2 at the optimal F1 threshold in Figure \ref{appendix:global-venn-diagram}. We observe that the majority of Cosine Similarity true positive identifications are also made by LSM-MS2, validate LSM-MS2's robustness with relation to a baseline ID workflow.

\begin{figure}[!ht]
    \centering
    \includegraphics[width=0.75\linewidth]{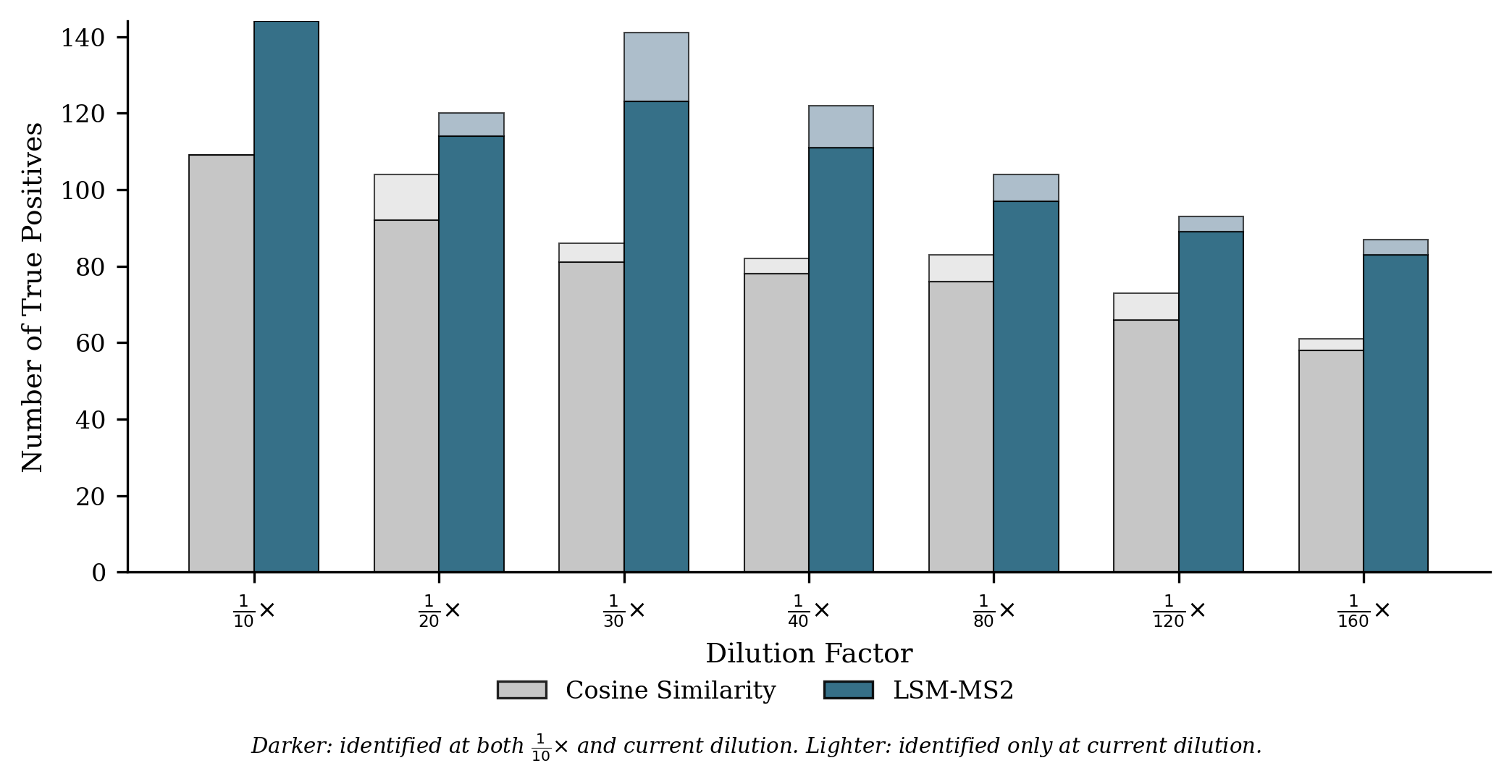}
    \caption{
    Consistency of true positive identifications across the NIST SRM1950 dilution series. 
    For each dilution factor, we compare the set of true positive analytes identified at that dilution to those detected at the highest concentration (1:10). 
    Shaded regions represent analytes consistently identified in both the 1:10 samples and the corresponding dilution, whereas lighter regions indicate analytes unique to that dilution. 
    }
    \label{appendix:dilution_consistency}
\end{figure}

To further assess the robustness of true positive identifications in the NIST dilution series, we compare the consistency of identified analytes across the dilution series. In theory, the analytes detected in each successive dilution should a subset of those acquired at high dilutions. While in practice the stochastic nature of MS/MS acquisition means that this ideal is not always achieved, this assumption can be used to provide a reasonable 2nd comparison of successful identification. Figure~\ref{appendix:dilution_consistency} shows the fraction of analytes identified both in the 1:10 (highest concentration) dilution and in each successively lower-concentration sample set. Most analytes detected by either Cosine Similarity or LSM-MS2 at any dilution were also present in the 1:10 sample. Within this consistently identified set, LSM-MS2 detected substantially more true positives than Cosine Similarity across all dilution levels. These results demonstrate that LSM-MS2 provides concentration-consistent identifications and maintains robustness to noise introduced at low-concentration regimes.

\subsection{Cosine Similarity Configuration and Ablation}

\begin{figure}[!ht]
  \centering
  \includegraphics[width=\textwidth]{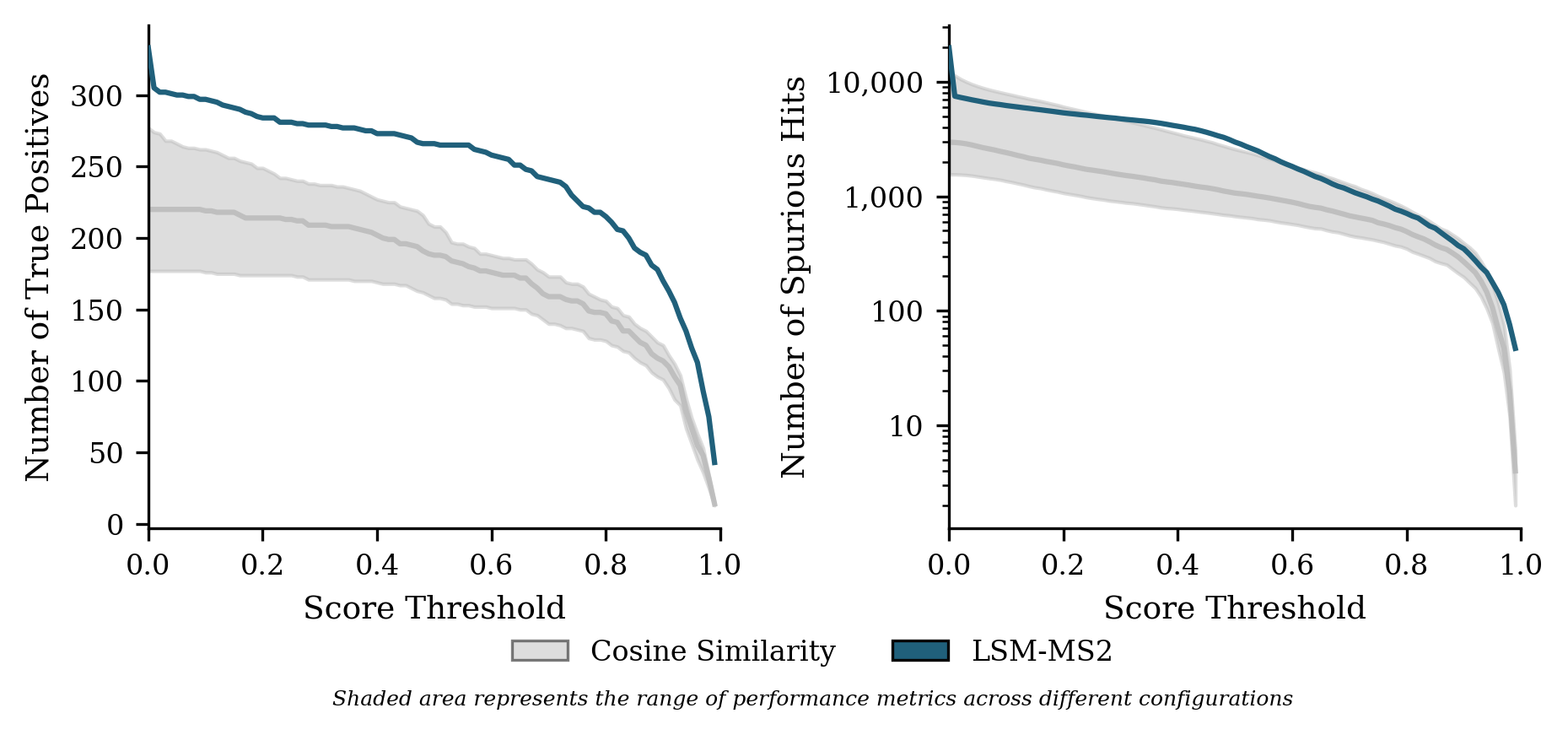}
  \caption{True positives and the number of spurious hits across score thresholds. Cosine Similarity is run five times with a minimum of 1–5 matched signals. Shaded areas represent variation across configurations, and the solid line represents median performance (n=3 matched signals).}
  \label{appendix:global-cross-n-matched}
\end{figure}

\begin{figure}[!ht]
    \centering
    \includegraphics[width=\textwidth]{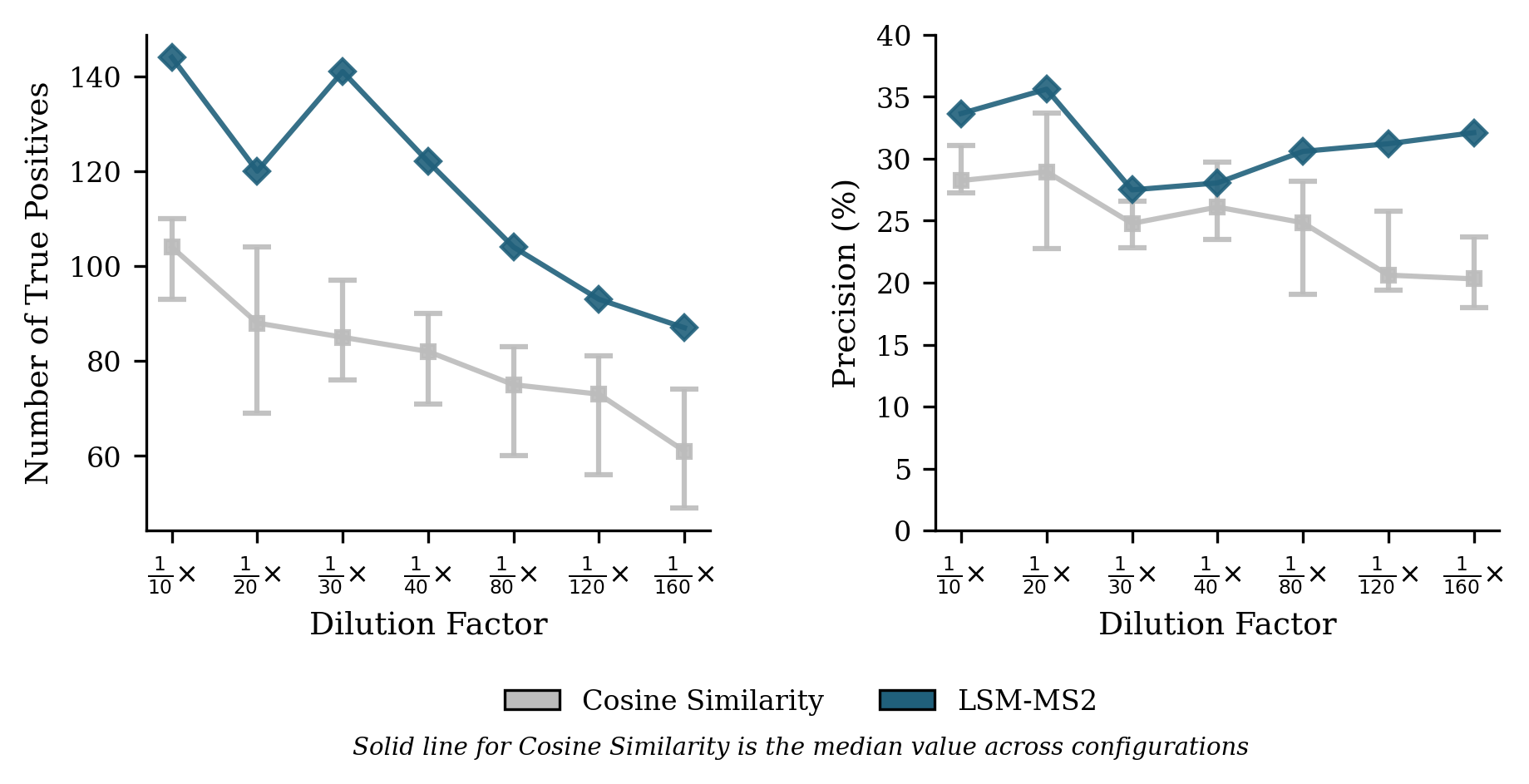}
    \caption{Number of true positives and precision stratified by dilution factor. Cosine similarity is run five times with a minimum of 1–5 matched signals, with the solid line representing the median value across configurations.}
    \label{appendix:n-matched-dilution}
\end{figure}

Finally, for our NIST dilution series, we calculated Cosine Similarity identifications using \textit{MZmine} \cite{MZmine}, a widely used analytical chemistry platform representing traditional expert-curated pipelines. To ensure a fair and algorithmically comparable baseline, we selected the Cosine Similarity configuration in MZmine that most closely matches our workflow. Unless otherwise noted, all figures use this configuration, with hyperparameters aligned to LSM-MS2: all scans were used for identification, a 10~ppm precursor tolerance, no precursor removal, a minimum of one matched signal, and weighted cosine similarity with square-root weighting as the similarity metric.  

We acknowledge that these parameter choices, particularly the minimum number of matched peaks, can substantially influence the number and quality of identifications. To assess this effect, we conducted an ablation study comparing LSM-MS2 against Cosine Similarity configurations using 1–5 minimum matched signals, as recommended by MZmine. Figure~\ref{appendix:global-cross-n-matched} shows the global number of true positives and spurious hits across score thresholds for each configuration. As expected, increasing the minimum matched signals reduces both true positives and spurious hits. Extending this analysis across dilution factors, we observe trends consistent with Figure~\ref{fig:dilution_optimal_f1}: LSM-MS2 consistently yields more true positive identifications with higher precision, with the precision gap larf at higher dilution levels.  

While we recognize that alternative MZmine parameterizations may shift absolute results, we chose to evaluate using the configuration most directly comparable to our workflow to isolate differences attributable to the identification approach rather than parameter tuning.

\end{document}